\documentclass[runningheads]{llncs}

\usepackage[misc]{ifsym}
\usepackage{microtype}
\usepackage{graphicx}
\usepackage{booktabs}
\usepackage{dblfloatfix}
\usepackage{hyperref}
\usepackage{times}
\usepackage{epsfig}
\usepackage{graphicx}
\usepackage{amsmath}
\usepackage{amssymb}
\usepackage{makecell}
\usepackage[utf8]{inputenc} % allow utf-8 input
\usepackage[T1]{fontenc}    % use 8-bit T1 fonts
\usepackage{url}            % simple URL typesetting
\usepackage{booktabs}       % professional-quality tables
\usepackage{amsfonts}       % blackboard math symbols
\usepackage{nicefrac}       % compact symbols for 1/2, etc.
\usepackage{microtype}      % microtypography
\usepackage{wrapfig}
\usepackage{graphicx}
\usepackage{subfig}
\usepackage{caption}
\usepackage{comment}
\usepackage{xcolor}
\usepackage{amsmath}        % math symbols
\usepackage{algorithm}
\usepackage{algorithmic}
\usepackage{amsfonts, bm}   % Math font packages
\usepackage{eqparbox}
\usepackage[utf8]{inputenc} % allow utf-8 input
\usepackage[T1]{fontenc}    % use 8-bit T1 fonts
\usepackage{url}            % simple URL typesetting
\usepackage{booktabs}       % professional-quality tables
\usepackage{amsfonts}       % blackboard math symbols
\usepackage{nicefrac}       % compact symbols for 1/2, etc.
\usepackage{microtype}      % microtypography
\usepackage[color=gray!40, textsize=scriptsize]{todonotes}

\newcommand{\norm}[1]{\left\lVert#1\right\rVert}

\renewcommand\algorithmiccomment[1]{%
  \hfill\#\ \eqparbox{COMMENT}{#1}%
}
\author{Yunfei Teng\inst{1} \and Anna Choromanska\inst{1}\thanks{Senior lead.} \and Murray Campbell\inst{2} \and Songtao Lu \inst{2}\and Parikshit Ram \inst{2}\and Lior Horesh\inst{2}}
\authorrunning{Y.Teng et al.}
\institute{New York University, USA \and IBM Research, USA}

\begin{document}
\title{Overcoming Catastrophic Forgetting via Direction-Constrained Optimization}
\maketitle

\begin{abstract}
This paper studies a new design of the optimization algorithm for training deep learning models with a \textit{fixed} architecture of the classification network in a continual learning framework. The training data is non-stationary and the non-stationarity is imposed by a sequence of distinct tasks.
% This setting implies the existence of a manifold of network parameters that correspond to good performance of the network on all tasks. Our algorithm is derived from the geometrical properties of this manifold.
We first analyze a deep model trained on only one learning task in isolation and identify a region in network parameter space, where the model performance is close to the recovered optimum. We provide empirical evidence that this region resembles a cone that expands along the convergence direction. We study the principal directions of the trajectory of the optimizer after convergence and show that traveling along a few top principal directions can quickly bring the parameters outside the cone but this is not the case for the remaining directions. We argue that catastrophic forgetting in a continual learning setting can be alleviated when the parameters are constrained to stay within the intersection of the plausible cones of individual tasks that were so far encountered during training.
% Enforcing this is equivalent to preventing the parameters from moving along the top principal directions of convergence corresponding to the past tasks. 
Based on this observation we present our direction-constrained optimization (DCO) method, where for each task we introduce a linear autoencoder to approximate its corresponding top forbidden principal directions. They are then incorporated into the loss function in the form of a regularization term for the purpose of learning the coming tasks without forgetting. Furthermore, in order to control the memory growth as the number of tasks increases, we propose a memory-efficient version of our algorithm called compressed DCO (DCO-COMP) that allocates a memory of fixed size for storing all autoencoders. We empirically demonstrate that our algorithm performs favorably compared to other state-of-art regularization-based continual learning methods. The codes are publicly available at \url{https://github.com/yunfei-teng/DCO}.
\end{abstract}

\vspace{-0.3in}
\section{Introduction}\vspace{-0.1in}
\label{sec:intro}
% Humans are equipped with complex neurocognitive mechanisms that enable them 
A key characteristic feature of intelligence is the ability to continually learn over time by accommodating new knowledge and transferring knowledge between correlated tasks while retaining previously learned experiences. This ability is often referred to as \textit{continual or lifelong learning}. In a continual learning setting one needs to deal with a continual acquisition of incrementally available information from non-stationary data distributions (online learning) and avoid \textit{catastrophic forgetting}~\cite{mccloskey1989catastrophic}, i.e., a phenomenon that occurs when training a model on currently observed task leads to a rapid deterioration of the model's performance on previously learned tasks. In the commonly considered scenario of continual learning the tasks come sequentially and the model is not allowed to inspect again the samples from the tasks seen in the past~\cite{parisi2019continual}. Within this setting, there exist two types of approaches that are complementary and equally important in the context of solving the continual learning problem: i) methods that assume fixed architecture of deep model and focus on designing the training strategy that allows the model to learn many tasks and ii) methods that rely on existing training strategies (mostly SGD~\cite{bottou-98x} and its variants, which themselves suffer catastrophic forgetting~\cite{goodfellow2014empirical}) and focus on expanding the architecture of the network to accommodate new tasks. In this paper we focus on the first framework. 

Training a network in a continual learning setting, when the tasks arrive sequentially, requires solving many optimization problems, one per task. A space of solutions (i.e., network parameters) that correspond to good performance of the network on all encountered tasks determine a common manifold of plausible solutions for all these optimization problems. In this paper we seek to understand the geometric properties of this manifold. In particular we analyze how this manifold is changed by each new coming task and propose an optimization algorithm that efficiently searches through it to recover solutions that well-represent all previously-encountered tasks.
%\todo{``new coming task'' $\to$ ``new task'' or ``newly encountered task''?
%
%\textcolor{blue}{Yunfei}: I feel it is OK.
%}
% Our contribution to the existing literature relies on developing a continual learning algorithm that explicitly relies on the characteristics of the manifold shared between tasks. What is new in this paper?  To the best of our knowledge, the analysis of the deep learning loss landscape that determines the shape of this manifold, the algorithm, and the experimental results are all new here.
The contributions of our work could be summarized as follows:
\begin{itemize}
\item We empirically analyse the deep learning loss landscape and show that there is a cone in the network parameter space where the model performance is close to the recovered optimum.
%\todo{``make an empirical analysis'' $\to$ ``empirically analyse''?}
%\todo{maybe clarify that we are talking about model performance on already learned task
%
%\textcolor{blue}{Yunfei}: Not really. It is a more general conclusion.
%}
\item We propose a new regularization-based continual learning algorithm that explicitly encourages the model parameters to stay inside the plausible cone by identifying a few top forbidden principal directions for each task.
\item We propose an autoencoder architecture that significantly reduces the memory complexity to save the top forbidden principal directions for a given task.
\item We design a compression method to control the memory growth and avoid introducing a new autoencoder per task, thereby requiring only a constant size memory overhead irrespective of the number of tasks.
%\todo{maybe say ``new autoencoder per task, thereby requiring only a constant size memory overhead irrespective of the number of tasks''}
\end{itemize}

The paper is organized as follows: Section~\ref{sec: related_work} reviews recent progress in the research area of continual learning, Section~\ref{sec:observation} provides empirical analysis of the geometric properties of the deep learning loss landscape and builds their relation to the continual learning problem, Section~\ref{sec:algorithm} introduces our algorithm that we call DCO since it is based on the idea of direction-constrained optimization, Section~\ref{sec:exp_res} contains empirical evaluations, and finally Section~\ref{sec:conclusion} concludes the paper. Additional results are contained in the Supplement.\vspace{-0.12in}

\section{Related Work}\vspace{-0.12in}
\label{sec: related_work}
Continual learning and the catastrophic forgetting problem has been addressed in a variety of papers. A convenient literature survey dedicated to this research theme was recently published~\cite{parisi2019continual}. The existing approaches can be divided into three categories \cite{parisi2019continual,robust_evaluation_cl} listed below.

\textbf{Regularization-based methods} modify the objective function by adding a penalty term that controls the change of model parameters when a new task is observed. In particular these methods ensure that when the model is being trained on a new task, the parameters stay close to the ones learned on the tasks seen so far. EWC~\cite{EWC} approximates the posterior of the model parameters after each task with a Gaussian distribution and uses tasks' Fisher information matrices to measure the overlap of tasks. The idea is extended in~\cite{ritter2018online} where the authors introduce Kronecker factored Laplace approximation. SI~\cite{Synapses} introduces the notion of synaptic importance, enabling the assessment of the importance of network parameters when learning sequences of classification tasks, and penalizes performing changes to the parameters with high importance when training on a new task in order to avoid overwriting old memories. Relying on the importance of the parameters of a neural network when learning a new task is also a characteristic feature of another continual learning technique called MAS~\cite{MAS}. The RWALK method~\cite{RWALK} is a combination of an efficient variant of EWC and a modified SI technique that computes a parameter importance score based on the sensitivity of the loss over the movement on the Riemannian manifold. Additionally, RWALK stores a small subset of representative samples from the previous tasks and uses them while training the current task, which is essentially a form of a replay strategy described later in this section. The recently proposed OGD algorithm~\cite{OGD} and its variant GPM~\cite{saha2021gradient} rely on constraining the parameters of the network to move within the orthogonal space to the gradients of previous tasks. \cite{OGD} is memory-consuming and not scalable as it requires saving the gradient directions of the neural network predictions on previous tasks.
Finally, the recursive method of~\cite{liu2022continual} modifies the gradient direction of each step to minimize the expected forgetting by introducing an additional projection matrix which requires a per-step update with linear memory complexity in the number of model parameters. All methods discussed so far constitute a family of techniques that keep the architecture of the network fixed. The algorithm we propose in this paper also belongs to this family.

Another regularization method called LwF~\cite{LwF} optimizes the network both for high accuracy on the next task and for preservation of responses on the network outputs corresponding to the past tasks. This is done using only examples for the next task. The encoder-based lifelong learning technique~\cite{encoder_cl} uses per-task under-complete autonecoders to constraint the features from changing when the new task arrives, which has the effect of preserving the information on which the previous tasks are mainly relying. Both these methods fundamentally differ from the aforementioned techniques and the approach we propose in this paper in that they require a separate network output for each task. Finally, P\&C~\cite{schwarz2018progress} builds upon EWC and takes advantage of the knowledge distillation mechanism to preserve and compress the knowledge obtained from the previous tasks. Such a mechanism could as well be incorporated on the top of SI, MAS, or our technique.
%\todo{what are ``features'' in ``.. constraint the features from changing ...''? is it network parameters?
%
%\textcolor{blue}{Yunfei}: ``feature'' is a very specific concept introduced in that paper and thus I would not elaborate more here.
%}
%\todo{what is meant by ``separate network output for each task'' and how is that different from us using a AE for each task in vanilla DCO? I think that can be elaborated.
%
%\textcolor{blue}{Yunfei}: We do have a discussion at the end of \S 4.2.
%}

The next two families of continual learning methods are not directly related to the setting considered in this paper and are therefore reviewed only briefly.

\textbf{Dynamic architecture methods} either expand the model architecture~\cite{aljundi2017expert,rusu2016progressive,yoon2018lifelong,li2019learn} to allocate additional resources to accommodate new tasks (they are typically memory expensive) or exploit the network structure by parameter pruning or masking~\cite{mallya2018piggyback,mallya2018packnet}. Some techniques~\cite{hung2019compacting} interleaves the periods of network expansion with network compression, network pruning, and/or masking phases to better control the growth of the model. 

\textbf{Replay methods} are designed to train the model on a mixture of samples from a new task and samples from the previously seen tasks. The purpose of replaying old examples is to counter-act the forgetting process. Many replay methods rely on the design of sampling strategies~\cite{isele2018selective,aljundi2019gradient}. Other techniques, such as GEM~\cite{GEM}, A-GEM~\cite{chaudhry2018efficient} and MER~\cite{riemer2018learning}, use replay specifically to encourage positive transfer between the tasks (increasing the performance on preceding tasks when learning a new task). ORTHOG-SUBSPACE~\cite{chaudhry2020continual} reduces the interference between tasks by learning the tasks in different subspaces. Replay methods typically require large memory. Deep generative replay technique~\cite{shin2017continual,rostami2019complementary} addresses this problem and employs a generative model to learn a mixed data distribution of samples from both current and past tasks. Samples generated this way are used to support the training of a classifier. Finally, note that the setting considered in our paper does not rely on the replay mechanism.

In addition to the above discussed research directions, very recently authors started to look at task agnostic and multi-task continual learning where no information about task boundaries or task identity is given to the learner~\cite{NIPS2019_8981,he2019task,zeno2018task,Hou2019CVPR,Tao2020CVPR}. These approaches lie beyond the scope of this work.

\textit{Remark}: Regularization-based methods and replay methods are usually implemented with a fixed architecture of the classification network, but they require additional memory to save regularization terms or data samples. Conversely, dynamic architecture methods do not explicitly keep extra information in the memory, but they rely on the expansion and modification of the network architecture itself. Our approach falls into the family of regularization-based methods since we do not allow the architecture of the classification network to dynamically change and we also do not allow replay.
\vspace{-0.12in}

\section{Loss landscape properties}\vspace{-0.12in}
\label{sec:observation}%
The experimental observations provided in this section extend and complement the behavior characterization of SGD~\cite{feng2020neural} connecting its dynamics with random landscape theory that stems from physical systems. The results that will be presented here were obtained on MNIST and CIFAR-10 data sets (CIFAR-10 results are deferred to the Supplement). The details of the experimental setup of this section can be found in the Supplement (Section~\ref{app:observation_setup}).
%\todo{
%Is it necessary to discuss the architecture choice or cite some work that says that loss landscapes for SGD are qualitatively same across architectures? Otherwise won't this analysis depend on the architecture?\\
%\textcolor{red}{Anna:} we present two set of results on two different data sets, and I believe architectures, so I would not do that. Especially that we cannot really do this analysis for larger networks. So the fact that our algorithm works for a broad family of architectures and data sets, as shown in the Experimental section, is a proof that our observation seems to generalize across different architectures and data sets.\\
%\textcolor{cyan}{PR:} Thanks. Can we put your comments in the text or is it best to leave it as is?\\
%\textcolor{blue}{Yunfei:} The architectures are picked up from the other papers and people seem to use the similar architectures, so maybe we keep it for now.}%
Consider learning only one task. We analyze the top principal components of the trajectory of SGD \textit{after convergence}, i.e., after the optimizer reached a saturation level\footnote{The optimization process is typically terminated when the loss starts saturating but we argue that running the optimizer further gives benefits in the continual learning setting.}. Let $x^*$ denotes the value of the parameters in the beginning of the saturation phase. The convergence trajectory will be represented as a sequence of optimizer steps, where each step is represented by the change of model parameters that the optimizer induced (gradient). We consider $n$ steps after model convergence and compute the gradient of the loss function at these steps that we refer to as $\nabla L(x_1; \zeta_1),\nabla L(x_2; \zeta_2),\dots,\nabla L(x_n; \zeta_n)$ ($x_i$ denotes the model parameters at the $i^{\text{th}}$ step and $\zeta_i$ denotes the data mini-batch for which the gradient was computed at that step). We use them to form a matrix $G\in \mathbb{R}^{d \times n}$ ($i$-th column of the matrix is $\nabla L(x_i; \zeta_i)$) and obtain the eigenvectors $\{v_i\}$ of $GG^T$.\footnote{The explanation of the difference between $GG^T$ and the Fisher information matrix underlying the EWC method is deferred to the Supplement (Section \ref{app:efim}).} We furthermore define the averaged gradient direction $\bar{g} = \frac{1}{n}\sum_{i=1}^n \nabla L(x_i; \zeta_i)$. We first study the landscape of the deep learning loss function along directions $v_i$ and $\bar{g}$, i.e., we analyze the function
\begin{equation}
    \label{eq:ob1}
    f(\alpha, \beta, v_i) = L(x^* - \alpha\bar{g} + \beta v_i; \zeta),
\end{equation}
% \todo{can we motivate why we care about this $f(\alpha, \beta, v_i)$ function and what the following special cases imply in the context of CL?\\
% \textcolor{blue}{Yunfei:} we do explain the connection in the algorithm section \S 4.4 \\
% \textcolor{cyan}{PR:} Given the connection is so far away in the text, this might be confusing to the reviewers as to why we are looking at a function of this form. But if you have not had this issue in previous review cycles, we can leave it as is.\\
% \textcolor{blue}{Yunfei:} This is anyway a very helpful suggestion and I add one sentence to indicate we will explain the connection later.
% }
\noindent where $\alpha$ and $\beta$ are the step sizes along $-\bar{g}$ and $v_i$ respectively and $\zeta$ denotes the entire training data set. We will show how this function is connected to our algorithm later in section \ref{sec:res_algorithm}.

\textit{Remark}: Below, the eigenvector with the lower-index corresponds to a larger eigenvalue.

\begin{figure}[h!]\vspace{-0.1in}
    \subfloat[Fixing $\alpha=0$ and varying $\beta$.]{\label{fig:Dynamics2}
    \includegraphics[width=0.49\textwidth]{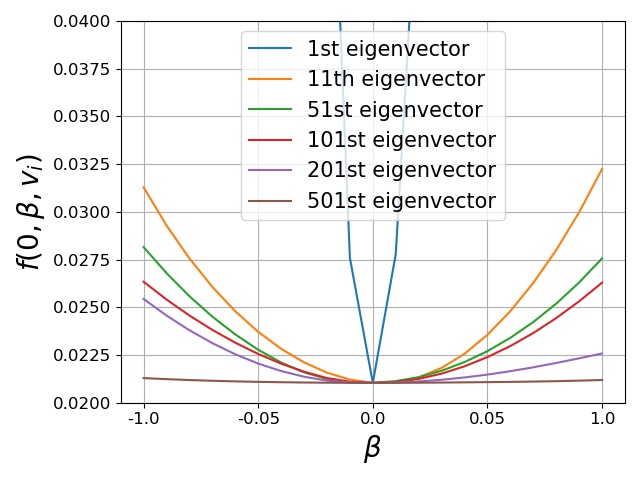}}
    \hfill
    \subfloat[Varying both $\sigma$ and $s$.]{\label{fig:landscape2}
    \includegraphics[width=0.49\textwidth]{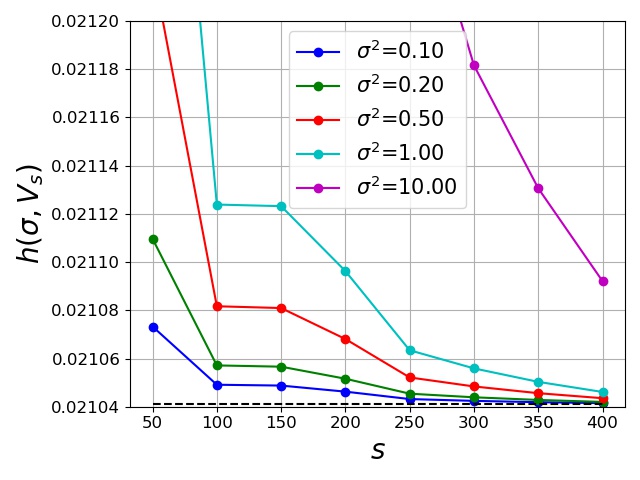}}
    \hfill
    \vspace{-0.1in}
    \caption{\textbf{left (a)}: The behavior of the loss function for $\alpha=0$ and varying $\beta$ when moving along different eigenvectors on MNIST (the complementary plot obtained on CIFAR-$10$ can be found in the Supplement, Fig. \ref{fig:ob1_cifar10}); \textbf{right (b)}: The behavior of the loss function when varying $\sigma$ and $s$ on MNIST (the complementary plot obtained on CIFAR-$10$ can be found in the Supplement, Fig. \ref{fig:ob2_cifar10}).}
    % \label{fig:landscape2} \label{fig:Dynamics2}
    \vspace{-0.25in}
\end{figure}

\textbf{Observation 1: Behavior of the loss for $\alpha=0$ and changing $\beta$} For each eigenvector $v_i$, we first fix $\alpha$ to $0$ and change $\beta$ in order to study the behavior of $f(0, \beta, v_i)$. Fig. \ref{fig:Dynamics2} captures the result. It can be observed that as the model parameters move away from the optimal point $x^*$ the loss gradually increases. At the same time, the rate of this increase depends on the eigendirection that is followed and grows faster while moving along eigenvectors with the lower-index. Thus we have empirically shown that \textit{the loss changes more slowly along the eigenvectors with the high index, i.e., the landscape is flatter along these directions.}
\vspace{-0.3in}

% \begin{figure}[h!]
%     \centering
%     \includegraphics[width=0.6\textwidth]{./toy_example/subspace/regularization_sigma_zoom.jpg}
%     \vspace{-0.2in}
%     \caption{The behavior of the loss function when varying $\sigma$ and $s$ on MNIST (the complementary plot obtained on CIFAR-$10$ can be found in the Supplement, Fig. \ref{fig:ob2_cifar10}).}
%     \vspace{-0.1in}
%     \label{fig:landscape2}
% \end{figure}

\begin{figure*}[t!]
 \vspace{-0.1in}
    \centering
    \includegraphics[width=0.32\linewidth]{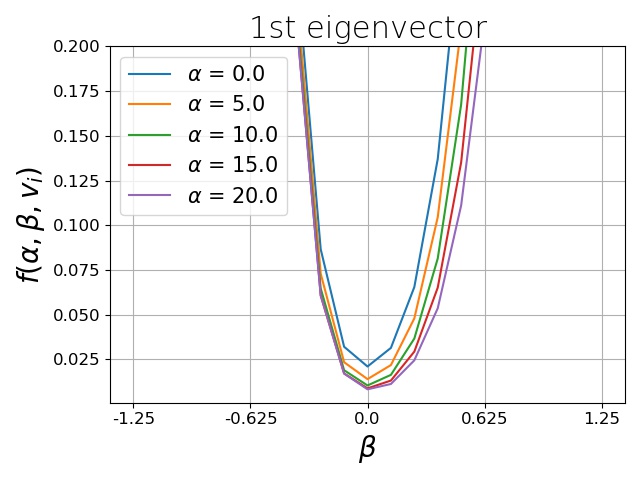}
    \includegraphics[width=0.32\linewidth]{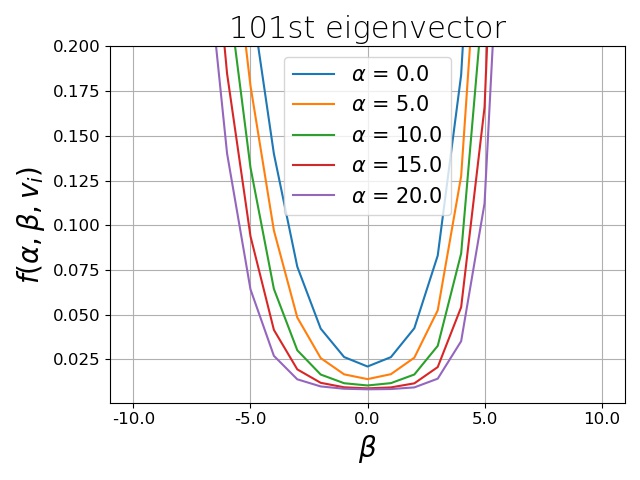}
    \includegraphics[width=0.32\linewidth]{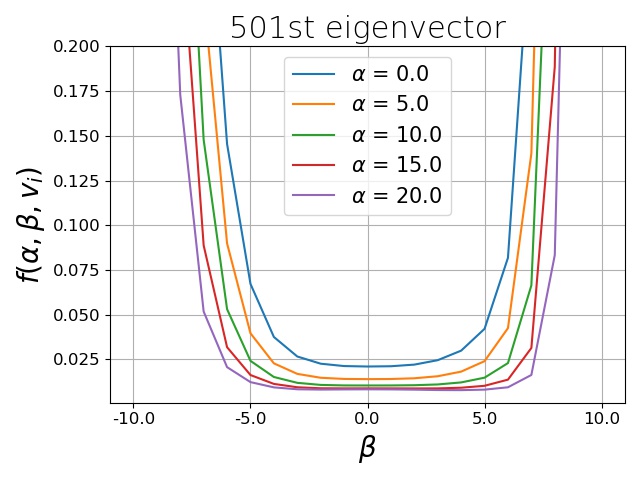}
    \vspace{-0.2in}
    \caption{The behavior of the loss function when both $\alpha$ and $\beta$ are changing for eigenvectors with different index on MNIST (the complementary plot obtained on CIFAR-$10$ can be found in the Supplement, Fig. \ref{fig:ob3_cifar10}).}
    \label{fig:landscape1}
    \vspace{-0.3in}
\end{figure*}

\textbf{Observation 2: Behavior of the loss in the subspaces spanned by groups of eigenvectors}
Here we generalize Observation 1 to the subspaces spanned by a set of eigenvectors. For the purpose of this observation only we consider the following metric instead of the one given in Equation~\ref{eq:ob1}:
\begin{equation}
\label{eq: projection}
h(\sigma, V_s) = \mathbb{E}_{\delta \sim \mathcal{N}(0, \frac{\sigma^2}{d}I)} L(x^* + V_sV_s^T \delta; \zeta),
\end{equation}
%\todo{Here we have not described what this new $f$ function is. Is it $L(\cdot, \zeta)$ from \eqref{eq:ob1}?
%
%\textcolor{blue}{Yunfei:} Yeah you are right.
%}
\noindent where $\delta$ is the random perturbation, $\sigma$ is the standard deviation, and $V_s = [v_{s-49}, v_{s-48}, \cdots, v_{s}]$ is the matrix of eigenvectors of $50$ consecutive indexes. To be more concrete, we locally (in the ball of radius $\sigma$ around $x^*$) sample the space spanned by the eigenvectors in $V_s$. The expectation is computed over $3000$ random draws of $\delta$. In Fig.~\ref{fig:landscape2} we examine the behavior of $h(\sigma,V_s)$ for various values of $\sigma$ and $s$. The plot confirms what was shown in Observation 1 that the loss landscape becomes flatter in the subspace spanned by the eigenvectors with high index.

\textbf{Observation 3: Behavior of the loss for changing $\alpha$ and $\beta$}
We generalize Observation 1 and examine what happens with $f(\alpha, \beta, v_i)$ when both $\alpha$ and $\beta$ change. Fig.~\ref{fig:landscape1} captures the result. We can see that as $\alpha$ increases, or in other words \textit{as we go further along the averaged gradient direction, the loss landscape becomes flatter.  This property holds for an eigenvector with an arbitrary index.} Thus for larger values of $\alpha$ we can go further along eigenvector directions without significantly changing the loss. This can be seen as a \textit{cone} that expands along $-\bar{g}$. Furthermore, the findings of Observation 1 are also confirmed in Fig.~\ref{fig:landscape1}. For the eigenvectors with higher index the loss changes less rapidly (the cone is wider along these directions). These properties underpin the design of new continual learning algorithm proposed in this work. When adding the second task, the algorithm constrains the optimizer to stay within the cone of the first task. Intuitively this can be done by first pushing the optimizer further into the cone along $-\bar{g}$ and then constraining the optimizer from moving along eigenvectors with low indexes in order to prevent forgetting the first task. This procedure can be generalized to an arbitrary number of tasks as will be shown in the next section.\vspace{-0.12in}

\section{Algorithm}\vspace{-0.12in}
\label{sec:algorithm}
In Section~\ref{sec:observation} we analyzed the loss landscape for a single task and discovered the existence of the cone in the model's parameter space where the model sustains good performance. We then discussed the consequence of this observation in the continual learning setting. In this section we propose a \textit{tractable} continual learning algorithm that for each task finds its cone and uses it to constrain the optimization problem of learning the following tasks. We refer to the model that is trained in the continual learning setting as $\mathcal{M}$. The proposed algorithm relies on identifying the top directions along which the loss function for a given task increases rapidly and then constraining the optimization from moving along these directions (we will refer to these directions as ``prohibited'') when learning subsequent tasks. Note that each new task adds prohibited directions. In order to efficiently identify and constrain the prohibited directions we use \textit{reduced linear autoencoders} whose design was tailored for the purpose of the proposed algorithm. We train separate autoencoders for each learned task. The $j^{\text{th}}$ autoencoder admits on its input gradients of the loss function that are obtained when training the model $\mathcal{M}$ on the $j^{\text{th}}$ task. The intuitive idea behind this approach is that autoencoder with small feature vector will capture the top directions of the gradients it is trained on. We refer to our method as \textit{direction-constrained optimization} (DCO) method. Furthermore, we will show that we can relax the need to allocate new memory for each task and propose a memory-efficient version of our algorithm called \textit{compressed} DCO (DCO-COMP) which requires a memory of fixed size for storing all autoencoders.
%todo{Mention here that we will relax the need for per-task AEs with the compressed one}

\subsection{Loss function}
We next explain the loss function that is used to train the model $\mathcal{M}$ in a continual learning setting. 
From Section \ref{sec:observation}, we recognize that the matrix $G$ formed by the gradients obtained from the current task can be used to describe the properties of the loss landscape. Furthermore, we can incorporate it as a regularization term into the loss function to prevent the increment of loss for the current task when training for a future task.
Therefore, the loss function that is used to train the model on the $i^{\text{th}}$ task takes the form:
\begin{equation}
\label{eq:general_obj}
L_i (x;\xi) = L_{ce}(x;\xi) + \lambda \sum_{j=1}^{i-1}\norm{(G^j)^T(x-x_j^*)}_2^2,
\end{equation}
where $\xi$ is a training example, $L_{ce}$ is a cross-entropy loss, $\lambda$ is a hyperparameter controlling the strength of the regularization, $G^j$ is the regularization matrix whose columns are the sampled gradients, and $x_j^*$ are the parameters of model $\mathcal{M}$ obtained at the end of training the model on the $j^{\text{th}}$ task. However, directly saving matrix $G^j$ will make the algorithm become \textit{intractable}. Thus, we instead introduce an autoencoder $ENC^j$ to approximate Eq. \ref{eq:general_obj} by:

\begin{equation}
\label{eq:main_obj}
L_i (x;\xi) = L_{ce}(x;\xi) + \lambda \sum_{j=1}^{i-1}\norm{ENC^j(x-x_j^*)}_2^2,
\end{equation}
where $ENC^j(\cdot)$ is the operation of the encoder of the autoencoder trained on task $j$. The linear autoencoders with appropriate regularization are able to recover the principal components of gradients~\cite{NEURIPS2020_4dd9cec1}. The top principal components correspond to the directions where the loss changes the quickest and we consider these directions as the prohibited directions. This is well-aligned with our observations from Section \ref{sec:observation}.

\begin{algorithm} %Complete Continual Learning algorithm
\caption{DCO/DCO-COMP Algorithm} % \caption should be before \label
\label{alg:continual_learning}
\begin{algorithmic}
\REQUIRE $\eta$ and $\eta_a$: learning rates of the model and autoencoders respectively. $\gamma_1, \gamma_2 \in (0,1]$: pulling strengths that controls the searching scope of the model parameters. $N$: number of additional epochs used to train the model after saturation. 
$C$: number of points to average. $\theta$: step size for pushing the optimizer inside the cone ($\theta\leq1$ corresponds to parameter interpolation; $\theta>1$  corresponds to parameter extrapolation). 
$m$: the size of the batch of gradients fed into autoencoders. $\tau$: the period of updates of the model parameters in step 3. $n$: number of tasks. $\mathcal{T}=\{\mathcal{T}_1, \dots, \mathcal{T}_n\}$: training data from task $1,2,\dots, n$. $|\mathcal{T}_i|$: number of iterations (mini-batches) required to process all data samples from task $i$.
\STATE
\STATE \hspace*{-1.25em} \textbf{Procedure:}
\FOR{$i=1$ {\bfseries to} $n$}
\STATE \# \underline{step 1: train model until convergence}
\STATE $x_0 \leftarrow x$
\REPEAT
\STATE $\xi \leftarrow$ randomly sample from $\mathcal{T}_i$
\STATE $x \leftarrow x - \eta \nabla_x L_i(x;\xi) - \gamma_1 (x - x_0)$ 
\UNTIL convergence

\STATE
\STATE \# \underline{step 2: push the model parameters into the cone}
\STATE $x_1 \leftarrow 0, x_2 \leftarrow 0$
\FOR{$j=1$ {\bfseries to} $N \times |\mathcal{T}_i|$}
\STATE $\xi \leftarrow$ randomly sample from $\mathcal{T}_i$
\STATE $x \leftarrow x - \eta \nabla_x L_{ce}(x;\xi)$
\STATE \bf{if} $j \leq C$ \bf{then} $x_1 \leftarrow x_1 + x$ 
\STATE \bf{else if} $j > N \times |\mathcal{T}_i| - C$ \bf{then} $x_2 \leftarrow x_2 + x$
\ENDFOR
\STATE $x^*_i \leftarrow x_1 - \theta\frac{(x_1-x_2)}{\norm{x_1-x_2}}$ \{push into the cone\}

\STATE
\STATE \# \underline{step 3: train autoencoder until convergence}
\REPEAT
\STATE $g \leftarrow 0, G \leftarrow \{\}$
\FOR{$j=1$ {\bfseries to} $m$}
\STATE $\xi \leftarrow$ randomly sample from $\mathcal{T}_i$
\STATE $g \leftarrow g + \nabla_x L_{ce}(x;\xi)$; $G \leftarrow G \cup \nabla_x L_{ce}(x;\xi)$
\STATE \bf{if} {$\tau$ divides $j$} \bf{then} $x \leftarrow x - \eta g$; $g \leftarrow 0$
\ENDFOR
\STATE $G \leftarrow \frac{G}{\sqrt{\norm{G}_2^2/ m}}$ \{Normalize batch of gradients\}
\STATE $W \leftarrow W - \frac{\eta_a}{m} \nabla_W L_{mse}(W;G)$; $x \leftarrow x - \gamma_2 (x - x_i^*)$
\UNTIL convergence
\STATE

\STATE \# \underline{step 4: store autoencoder parameters}
\STATE $W^i \leftarrow W$ 
% \STATE $x \leftarrow x_i^*$ \{Restore model parameters\}
\IF{use DCO-COMP}
\STATE compress and update $\{W^1, \cdots, W^i\}$ by Equation \ref{eq:dco_comp1} and Equation \ref{eq:dco_comp2}.
\ENDIF
\ENDFOR
\STATE
\STATE \hspace*{-1.25em} \textbf{Output:} $x_n^*$
\end{algorithmic}
\end{algorithm}

\subsection{Reduced linear autoencoders}
\label{sec:rla}
In our algorithm, the role of autoencoder is to identify the top $k$ directions of the optimizer's trajectory after convergence, where this trajectory is defined by gradient steps, obtained during training the model $\mathcal{M}$. A traditional linear autoencoder, consisting of two linear layers, would require $2 \times d \times k$ number of parameters, where $d$ denotes the number of parameters of the model $\mathcal{M}$. Commonly used deep learning models however contain millions of parameters \cite{NIPS2012_4824,Simonyan15,He2016DeepRL}, which makes a traditional autoencoder not tractable for this application. In order to reduce the memory footprint of the autoencoder we propose an architecture that is inspired by the singular value decomposition. The proposed autoencoder admits a matrix on its input and is formulated as
\begin{equation} 
 AE(M)=Udiag(U^\top MV)V^\top,
\end{equation}
where $diag(U^\top MV)$ is a matrix formed by zeroing out the non-diagonal elements of $U^\top MV$, $M$ is an autoencoder input matrix of size $m \times n$, and $U$ and $V$ are autoencoder parameters of size $m \times k$ and $n \times k$ respectively. Thus, the total number of parameters of the proposed autoencoder is $k(n + m)$, which is significantly lower than in case of traditional autoencoder ($knm$), especially when $n$ and $m$ are large. We call this architecture a \textit{reduced linear autoencoder}.

We use a separate encoder $ENC_l$ and decoder $DEC_l$ for each layer $l$ of the model $\mathcal{M}$. We couple them between layers using a common ``feature vector'' which is created by summing outputs of all encoders. This way the feature vector will contain the information from all layers. The proposed autoencoder is then formulated as
\begin{eqnarray}
    AE(G) &=& \{DEC_1(ENC(G)), \nonumber \\ 
    &\dots,&DEC_L(ENC(G))\},\\
    \text{where} \nonumber\\
    ENC(G) &=& \sum_lENC_l(G_l), \\
    ENC_l(G_l) &=& diag(U_l^\top G_l V_l), \\
    DEC_l(ENC(G)) &=& U_l ENC(G) V_l^\top,
\end{eqnarray}
$G=\{G_1, G_2, \dots, G_L\}$ is a set of matrices such that each matrix contains gradients of the model for a given layer, and $L$ is number of layers in the model. Finally, in order to enable processing the gradients of the convolutional layers we reshape them from their original size $o\times i \times w \times h$ to $o \times iwh$, where $o$ is number of output channels, $i$ is number of input channels, and $w$ and $h$ are width and height of the kernel of the convolutional layer. We train the autoencoder with standard mean square error loss
\begin{eqnarray}
    L_{mse}(W;G)=\norm{AE(G) - G}_2^2,
\end{eqnarray}
where $W=\{U_1, V_1, \dots, U_L, V_L\}$ is set of autoencoder's parameters. In the next section we propose a memory efficient variant DCO-COMP and show a compression scheme which allows us to avoid scaling the memory size as the number of tasks increases and results in a solution with a fixed memory size.

\textit{Remark:} Using one autoencoder per task in a continual learning context has been explored in the literature before. For example,~\cite{encoder_cl} uses an autoencoder to capture the features that are crucial for its corresponding task. Authors show experiments for only two tasks. Another method~\cite{aljundi2017expert} embeds autoencoders into the classification network to identify the tasks and make predictions. How do we differ from these approaches? First, we utilize autoencoders to encode optimizer directions. Second, as opposed to~\cite{aljundi2017expert} the autoencoders are not used within the classification network, thus they are not utilized at testing, but only at training. Third, as opposed to~\cite{encoder_cl} we demonstrate experiments on multiple tasks. Finally, note that autoencoders have not been used before to support parameter-wise regularization-based continual learning frameworks.

%Using autoencoders results in a memory overhead compared to existing autonencoder-free regularization-based frameworks such as~\cite{EWC,Synapses,RWALK} (our memory consumption is about ten times larger than these methods). At the same time using the autoencoders gives us a large reduction of memory complexity (two orders of magnitude) compared to methods like~\cite{OGD}. The computations involving autoencoders in our approach can also be easily parallelized. This is not the case for other methods such as~\cite{Synapses,RWALK} that rely on computing the change in loss over an entire trajectory through the parameter space. Nevertheless, in the next section we propose a way to eliminate the memory problem of DCO by suggesting a compression scheme which allows us to avoid scaling the memory size as the number of tasks increases and results in a solution with a fixed memory size.

\vspace{-0.15in}
\subsection{Compression of autoencoders}
\label{sec: dco_comp}
\begin{figure*}[t!]
\centering
\includegraphics[width=0.95\textwidth]{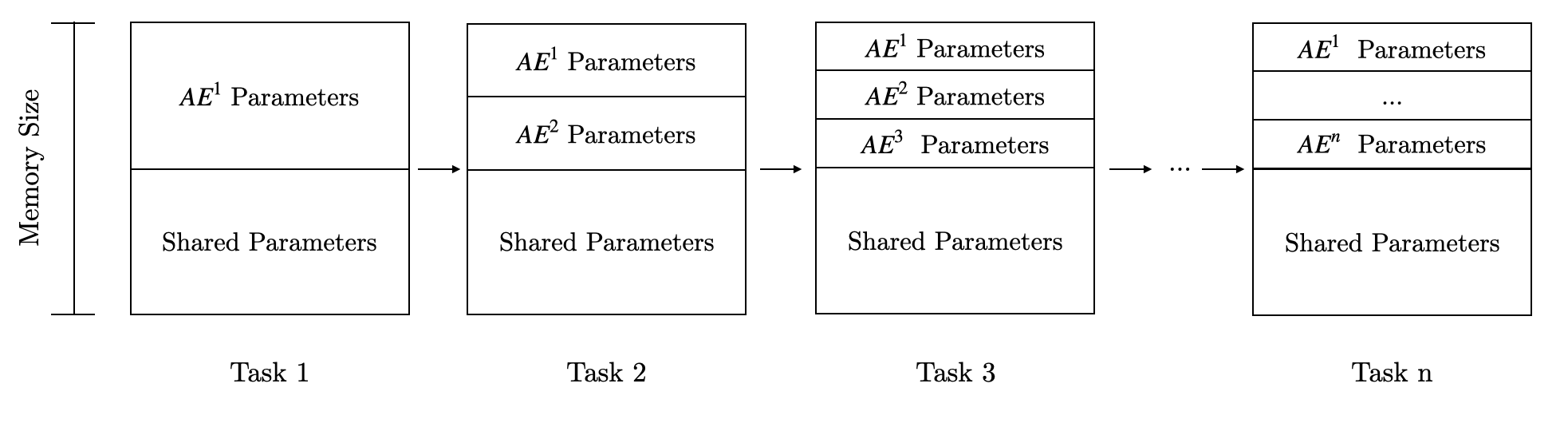}
\vspace{-0.2in}
\caption{The memory size of the autoencoders remains unchanged across the tasks.}
\label{fig:shared_autoencoder2}
\vspace{-0.2in}
\end{figure*}

% control the memory growth as the number of tasks increases
To avoid scaling the memory size as the number of tasks increases, we compress the autoencoders recursively so that only a \textit{constant} memory of size $k \times (m+n)$ is required to store all autoencoders during training. More specifically, after training on the $i^{th}$ task, all autoencoders are compressed together such that each autoencoder keeps $\frac{1}{2i} \times k \times (m+n)$ parameters separately. On top of that, by introducing shared parameters across autoencoders, whose size is fixed to $\frac{1}{2} \times k \times (m+n)$, we ensure that the information that would be lost due to compression is instead partially absorbed by the shared parameters. The memory allocation of autoencoders on each task is illustrated in Fig \ref{fig:shared_autoencoder2}.
% \todo{\textcolor{cyan}{PR:} Is this $k$ connected to the number of prohibited directions we are considering? If yes, we say above that we are using the reduced AE of size $k(n +m)$ per layer. So how does $d$ come in the picture since we earlier argue that $kd$ is too large?

% \textcolor{blue}{Yunfei}: Your are right. This is a critical typo.
% }
% With this setting, the total number of parameters for autoencoder is fixed to $k \times d$ through the whole training process. 
% After training on $i^{th}$ task, we reduce the sizes of all the autoencoders altogether so that each autoencoder maintains an individual set of parameters of size $\frac{k}{2i} \times d$. Furthermore, we assume these autoencoders could share a part of information in common and thus a set of parameters of size $\frac{k}{2} \times d$ are shared across all the autoencoders.
% Besides, the information lost through compression process will be absorbed into the shared autoencoder. 

Denote the shared parameters and the $j^{th}$ compressed autoencoder's parameters as $\bar{W} = \{\bar{U}_1, \bar{V}_1, \dots, \bar{U}_L, \bar{V}_L\}$ and $\tilde{W}^j=\{\tilde{U}_1^j, \tilde{V}_1^j, \dots, \tilde{U}_L^j, \tilde{V}_L^j\}$, respectively. Similar as before, we use a separate encoder $\widehat{ENC}_l$ and decoder $\widehat{DEC}_l$ for each layer $l$. The formulation of the $j^{th}$ compressed autoencoder is given as
% On the $i^{th}$ task we compress each autoencoder (including the shared autoencoder) from past by
\begin{eqnarray}
\widehat{AE}^j(W^j) &=& \{\widehat{DEC}^j_1(\widehat{ENC}^j(W^j)), \nonumber \\ 
&\dots,&\widehat{DEC}^j_L(\widehat{ENC}^j(W^j))\},\\
\text{where} \nonumber\\
\widehat{ENC}^j(W^j_l) &=& \sum_l \widehat{ENC}^j_l(W_l^j), \\
\widehat{ENC}^j_l(W^j_l) &=& diag\left(\bar{U}_l^\top U_l^j (V_l^j)^\top\bar{V}_l + (\tilde{U}_l^j)^\top U_l^j (V_l^j)^\top\tilde{V}_l^j\right), \\
\widehat{DEC}^j_l(\widehat{ENC}^j(W^j)) &=& \tilde{U}_l^j \widehat{ENC}^j(W^j) (\tilde{V}_l^j)^\top
% ENC'(W^j) &=& diag \left(\sum_{l=1}^L \bar{U}_l^\topU_l^j (V_l^j)^\top\bar{V}_l + \sum_{l=1}^L (\tilde{U}_l^j)^\topU_l^j (V_l^j)^\top\tilde{V}_l^j \right)\\
% ENC_l^{i}(W^i; W^j) = diag \left(\sum_{l=1} (U^i_l)^\topU_l^j (V_l^j)^\topV^i_l \right), \\
% DEC'_l(ENC(W^j)) &=& \tilde{U}_l^j ENC(W^j) (\tilde{V}_l^j)^\top
% \sum_{j=1}^i   D(\bar{W}; W^j)   - U_l^j (V_l^j)^T
\end{eqnarray}
% \todo{The math above needs to be fixed. In \S 4.2, the input to the AE was $G$ but here the input to the AE are the AE parameters $W, \bar W, W^j$ themselves which is confusing\\
% \textcolor{blue}{Yunfei}: this is something expected. In \S 4.2, we are training the autoencoders on the gradients; However, when we compress them altogether, we do not have the saved gradients anymore and thus we make compression directly on the autoencoder parameters.\\
% \textcolor{cyan}{PR:} Some notation might still need (minor) fixes. We define $W^j=\{\tilde U_l^j, l \in [L] \}$ but we dont define (i)~$\tilde W^j$ used in (15), (ii)~$W_l^j$ in (12) and $U_l^j, V_l^j$ used in (13). We only define them later in (16). Also in (14) I think we need $\widehat{ENC}^j$ on the right hand side.\\
% \textcolor{blue}{Yunfei}:You are right. These notations have been fixed now.
% }
\noindent where $W^j=\{U_1^j, V_1^j, \dots, U_L^j, V_L^j\}$ is set of $j^{th}$ autoencoder's uncompressed parameters. We train the compressed autoencoders with standard mean square error loss:
\begin{equation}
\label{eq:dco_comp1}
L_{mse}(\tilde{W}^1, \cdots, \tilde{W}^i, \bar{W}; W^1,\cdots, W^i) = \sum_{j=1}^i \norm{\widehat{AE}^j(W^j) -W^j}_2^2
\end{equation}
Then we assign the $j^{th}$ autoencoder with a new set of parameters:
\begin{align}
\label{eq:dco_comp2}
W^j 
% &= \{U_1^j, V_1^j, \cdots, U_L^j, V_L^j\} \nonumber\\
&= \{(\bar{U}_1, \tilde{U}_1^j), (\bar{V}_1, \tilde{V}_1^j), \cdots,(\bar{U}_L, \tilde{U}_L^j),(\bar{V}_L, \tilde{V}_L^j)\}
\end{align}
\textit{Remark}: We are not able to re-access the model gradients from previous tasks anymore, but the learned prohibited directions for each task could be recovered from the corresponding autoencoder parameters. Thus we make compression directly on the autoencoder parameters.
% \begin{figure*}
% \centering 
% \includegraphics[width=0.95\textwidth]{autoencoders.png}
% \caption{Memory requirement remains constant as the number of tasks increases.}
% \label{fig:shared_autoencoder1}
% \end{figure*}
% \textcolor{red}{To distinguish from DCO, we name this memory efficient variant as compressed DCO (DCO-COMP).}

\subsection{Resulting algorithm}
\label{sec:res_algorithm}
The proposed algorithm comprises of four steps. In the first step we train the model $\mathcal{M}$ using the loss function proposed in Equation~\ref{eq:main_obj} until convergence. This loss function penalizes moving along ``prohibited'' directions recovered for the previous tasks. In the second step, we continue to train the model for additional $N$ epochs and make either interpolation or extrapolation (depending on the step size) between the averages of the first and the last $C$ points on the optimizer's trajectory. This step is equivalent to pushing the model parameters deeper into the cone and aligns well with Section~\ref{sec:observation} (see conclusions resulting from Fig.~\ref{fig:landscape1}). In the third step we train the autoencoder to recover ``prohibited'' directions for the current task. This again aligns well with Section~\ref{sec:observation} (see conclusions resulting from Fig.~\ref{fig:Dynamics2} and ~\ref{fig:landscape2}). Finally, we store the autoencoder parameters with or without compression. The algorithm's pseudo code is captured in Algorithm~\ref{alg:continual_learning}.\vspace{-0.12in}
% \todo{%Are these pulling strengths $\gamma_1, \gamma_2$ standard in CL or are they something we need to justify?
% %
% %Also, the update of $x$ within step 3, where are training the AE, seems a bit off. I thought that the model parameters $x$ is not longer updated once we push the model parameters into the cone.
% %
% \textcolor{blue}{Yunfei}: pulling strengths $\gamma_1, \gamma_2$ are not standard in CL but so far we have not seen any complains about that from the reviewers. The update of $x$ is required because we are sampling the gradients along the optimizer trajectory.\\
% \textcolor{cyan}{PR:} Sorry, I missed that this was in the loop. }

\section{Experiments}\vspace{-0.12in}
\label{sec:exp_res}
In this section we compare the performance of DCO and DCO-COMP with state-of-the-art regularization-based continual learning methods such as EWC \cite{EWC}, SI \cite{Synapses}, RWALK \cite{RWALK} and GPM~\cite{saha2021gradient}, as well as the vanilla SGD \cite{bottou-98x}. We use open source codes\footnote{https://github.com/facebookresearch/agem}\footnote{https://github.com/sahagobinda/GPM} for the experiments. 
% Note that A-GEM was proposed in a single-epoch setup originally. For a fair comparison, we run A-GEM for multiple epochs on training data as well. 
\vspace{-0.12in}

\subsection{Data sets and architectures} \vspace{-0.05in}
In our experiments we consider commonly used continual learning data sets: (1) \textbf{Permuted MNIST}. For each task we used a different permutation of the pixels of images from the original MNIST data set \cite{MNIST}. We generated $5$ data sets this way corresponding to $5$ tasks. (2) \textbf{Split MNIST}. We divide original MNIST data set into $5$ disjoint subsets corresponding to labels $\{\{0,1\}, \{2,3\}, \{4,5\}, \{6,7\}, \{8,9\}\}$. (3) \textbf{Split CIFAR-$\mathbf{100}$}. We divide original CIFAR-$100$ data set \cite{cifar} into $10$ disjoint subsets corresponding to labels $\{\{0-9\}, \cdots, \{90-99\}\}$. Additionally, we consider a cross-domain learning scenario, where tasks come from different domains. For a cross-domain learning experiment (\textbf{MNIST/Fashion MNIST}) we combine MNIST and Fashion MNIST data sets together.

For Permuted MNIST, Split MNIST, and MNIST/Fashion MNIST we use the original image of size $1 \times 28 \times 28$. We then normalize each image by mean $(0.1307)$ and standard deviation $(0.3081)$. For Split CIFAR-$100$, we use the original image of size $3 \times 32 \times 32$. We then normalize each image by mean $(0.5071, 0.4867, 0.4408)$ and standard deviation $(0.2675, 0.2565, 0.2761)$. Also, in the experiments with Split MNIST and Split CIFAR-$100$ we use a multi-head setup \cite{Synapses,RWALK} and we provide task descriptors \cite{GEM,chaudhry2018efficient} to the model at both training and testing.

Finally, for Permuted MNIST and MNIST/Fashion MNIST experiments we use a Multi-Layer Perceptron (MLP) with two hidden layers, each having $256$ units with ReLU activation functions (we refer to this architecture as MLP-$256$). For Split MNIST, we use a MLP with two hidden layers each having $100$ units with ReLU activation functions (we refer to this architecture as MLP-$100$). For Split CIFAR-$100$, we use the same convolutional neural network as in~\cite{RWALK} (we refer to this architecture as ConvNet). In all architectures we turn off the biases. 

\subsection{Training details}
To train the model, we use SGD optimizer \cite{bottou-98x} with momentum of $0.9$. The batch sizes are set to $128$, $128$, $128$ and $64$ respectively for MNIST/Fashion MNIST, Permuted MNIST, Split MNIST and Split CIFAR-$100$. 

For MNIST/Fashion MNIST, we use a constant learning rate of $1 \times 10^{-3}$. For Permuted MNIST and Split MNIST, we use a constant learning rate of $1 \times 10^{-3}$ and add weight decay penalty of $0.001$. For Split CIFAR-100, we use a learning rate of $1 \times 10^{-2}$ for the first task and then drop it by a factor of $0.1$ for the remaining tasks. 

For DCO on Split CIFAR-$100$, we also clip the $l_2$ norm of the gradients induced by regularization terms with a threshold of $1$ to avoid exploding gradient problem.

\subsection{Hyperparameters}

\begin{figure*}[htp!]
\vspace{-0.2in}
\centering
\includegraphics[width=.32\textwidth]{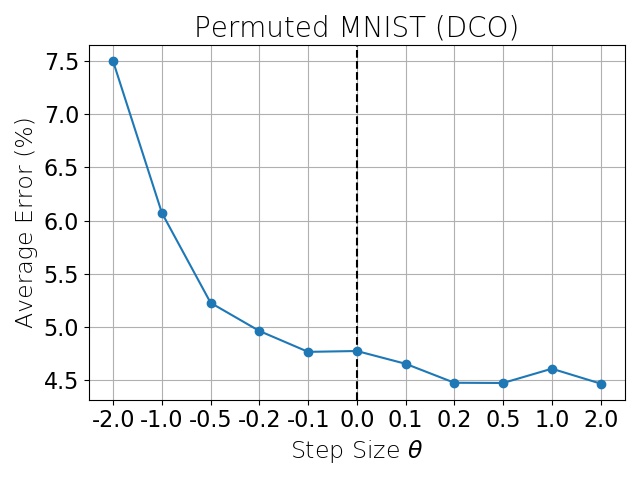}
\includegraphics[width=.32\linewidth]{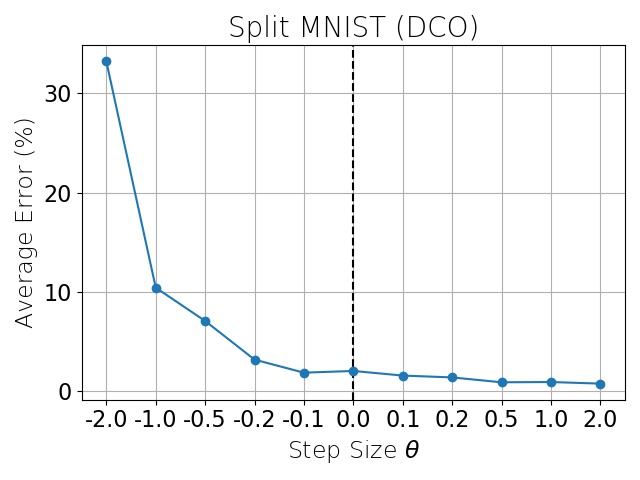}
\includegraphics[width=.32\linewidth]{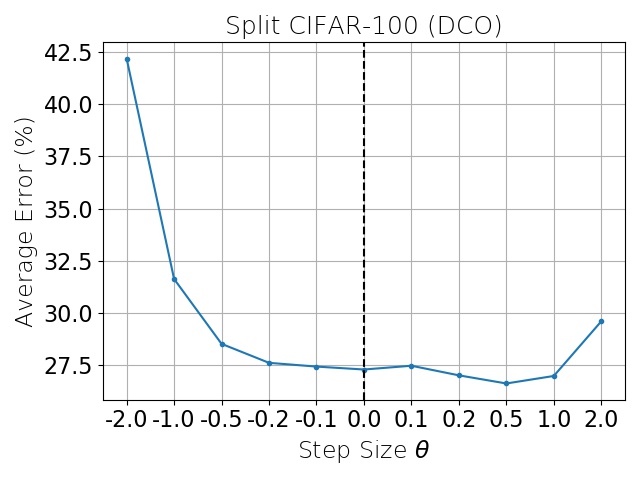}
\vspace{-0.15in}
\caption{Average error versus the step size $\theta$ (\textbf{left:} Permuted MNIST \textbf{middle:} Split MNIST \textbf{right:} Split CIFAR-$100$).}
\label{fig:pc_results}
\vspace{-0.2in}
\end{figure*}

The values of the hyperparameters explored in the experiments are reported in the Supplement (Section \ref{app:hyperparameters}). Here we illustrate how the final average error of DCO varies as the step size $\theta$ increases on Permuted MNIST, Split MNIST and Split CIFAR-100 (see Fig.~\ref{fig:pc_results}). The figure further confirms our findings in Section~\ref{sec:observation} that when moving deeper inside the cone, the performance improves. After some point the performance however eventually starts dropping, as can been seen on the right plot. This is most likely because of falling outside the cone, due to either inaccurate estimation of the direction pointing towards the center of the cone or the fact that the cone is bounded.

\subsection{Metric and Results}
We denote $e_{i,j}$ as \textit{test} classification error of the model on $j^{th}$ task after getting trained on $i$ tasks and evaluate the performance of each method based on the following metrics:
% \vspace{-0.2in}
\begin{enumerate}
\item \textbf{Average error}, which represents the average performance on all tasks learned so far. The average error $E_i^A$ on the $i^{th}$ task ($j \leq i$) is defined as $E_i^A = \frac{1}{i} \sum_{j=1}^i e_{i, j}$. 
\item \textbf{Forward interference (FWI) error}, which shows how preserving the knowledge of the previous tasks impairs the model's learning ability for a new task. The final FWI error is defined as $E_n^{FWI} = \frac{1}{n}\sum_{j=1}^n e_{j,j}$.
% $E_n^{FWI} = 1-\frac{1}{n-1}\sum_{j=2}^n (e_{j-1,j} - e_{j,j})$.
\item \textbf{Backward transfer (BWT) error}~\cite{GEM}, which directly reflects how much the model has forgotten the previously learned tasks at the end of training. The final BWT error is defined as $E_n^{BWT} = \frac{1}{n}\sum_{j=1}^{n} e_{n,j} - e_{j,j}$.
% $E_n^{BWT} = \frac{1}{n-1}\sum_{j=1}^{n-1} e_{n,j} - e_{j,j}$.
\end{enumerate}
\vspace{-0.1in}

\begin{table}[H]
\vspace{-0.25in}
    \captionof{table}{Average Error $E^A_n$ (\%) for Permuted MNIST, Split MNIST, and Split CIFAR-100.}
    \vspace{-0.1in}
    \label{tab:results}
    \centering
    \begin{tabular}{|c||c|c|c|}
    \hline
    \bfseries{Method} & \bfseries{ Permuted MNIST } & \bfseries{ Split MNIST }& \bfseries{ Split CIFAR-100 }\\
    % & MNIST & MNIST & CIFAR-100 \\
    \hline
    SGD     & $17.41$ &  $9.68$  & $33.13$\\
    \hline
    EWC     & $7.29$  &  $2.7$   & $30.23$\\
    \hline
    SI      & $6.38$  & $2.29$   & $29.91$\\
    \hline
    RWALK   & $6.7$   & $5.67$    & $29.08$\\
    \hline
    % A-GEM   & $5.3$   & $2.36$    & $28.2$\\
    % \hline
    % OGD   &   &  &\\
    % \hline
    GPM     &$5.84$  & $4.97$    & $32.93$\\
    \hline
    \textbf{DCO} &$\mathbf{4.68}$ &$\mathbf{1.46}$ &$\mathbf{26.61}$\\
    \hline
    % \textbf{DCO-COMP}  &$\mathbf{4.84}$ &$\mathbf{1.34, 1.52, 1.60}$ &$\mathbf{28.22, 28.41, 28.45}$ \\
    \textbf{DCO-COMP}  &$\mathbf{4.84}$ &$\mathbf{1.34}$ &$\mathbf{28.22}$ \\
    \hline
    % \textbf{DCO-DIAG}  &$\mathbf{5.40}$ &$\mathbf{1.91}$ &$\mathbf{29.59}$\\
    % \hline
    \end{tabular}
\vspace{-0.2in}
\end{table}

\begin{figure*}
\centering 
\includegraphics[width=0.32\textwidth]{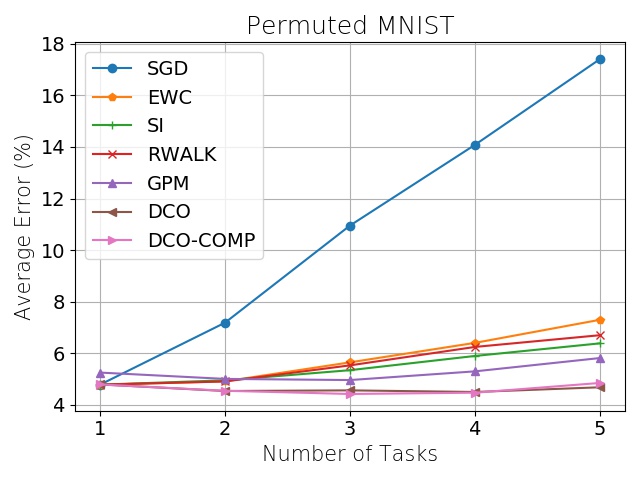}\hfill
\includegraphics[width=0.32\linewidth]{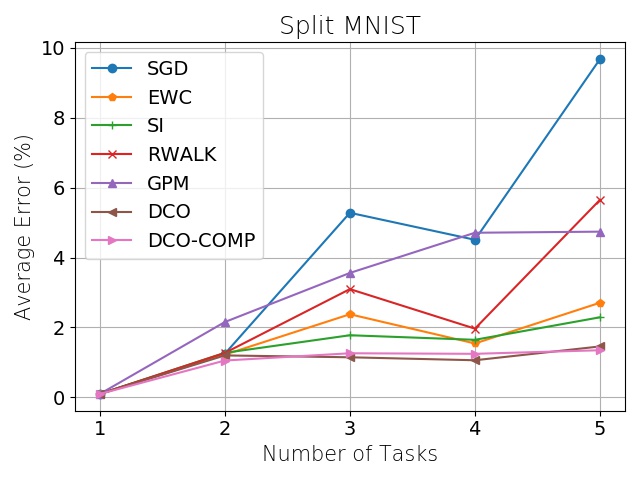}\hfill
\includegraphics[width=0.32\linewidth]{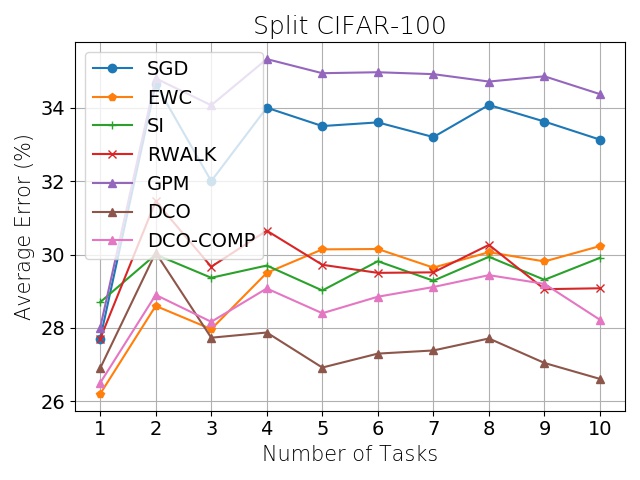}\hfill
\centering 
\includegraphics[width=0.32\textwidth]{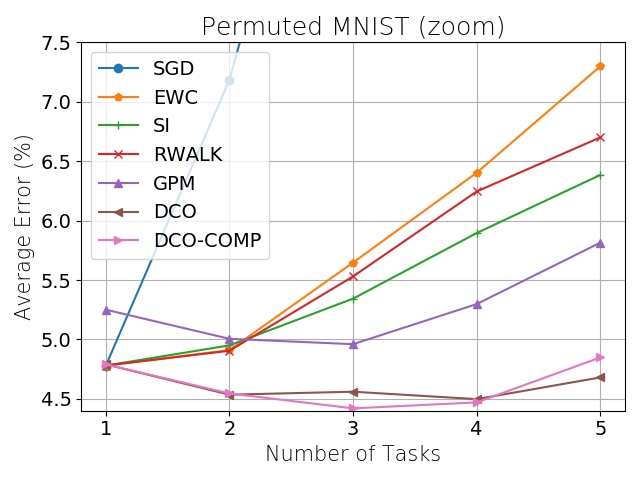}\hfill
\includegraphics[width=0.32\linewidth]{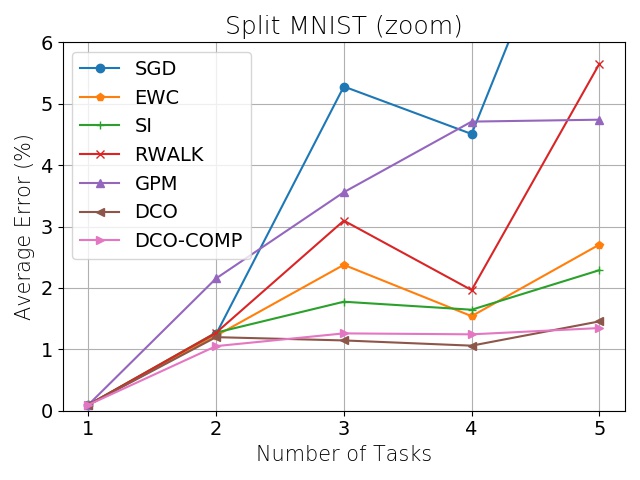}\hfill
\includegraphics[width=0.32\linewidth]{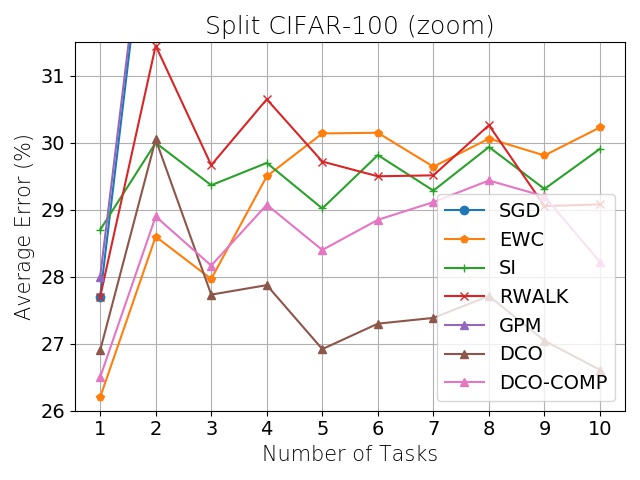}\hfill
\vspace{-0.15in}
\caption{Average error versus the number of tasks (original plots are on the top and zoomed are on the bottom; \textbf{left:} Permuted MNIST, \textbf{middle:} Split MNIST, \textbf{right:} Split CIFAR-$100$).}
\vspace{-0.45in}
\label{fig:results}
\end{figure*}

In Table \ref{tab:results} we demonstrate that DCO performs favorably compared to the baselines in terms of the final average error. In Fig.~\ref{fig:results} we show how the average error behaves when adding new tasks. The figure reveals that DCO consistently outperforms other methods. In most cases DCO-COMP performs similarly to DCO and across all experiments, just as DCO, it is superior to other techniques.
\vspace{-0.25in}

% \begin{figure*}
% \vspace{-0.1in}
% \centering 
% \includegraphics[width=0.32\textwidth]{main_experiments_sgd/permuted_mnist.jpg}
% \includegraphics[width=0.32\linewidth]{main_experiments_sgd/split_mnist.jpg}
% \includegraphics[width=0.32\linewidth]{main_experiments_sgd/split_cifar100.jpg}
% \vspace{-0.18in}
% \caption{Average error versus the number of tasks (the zoomed plots can be found in Supplement, Fig. \ref{fig:error_zoom}). \textbf{Left:} Permuted MNIST, \textbf{middle:} Split MNIST, \textbf{right:} Split CIFAR-$100$).}
% \vspace{-0.1in}
% \label{fig:results}
% \end{figure*}

% \begin{figure*}[h!]
% % \vspace{-0.15in}
% \begin{subfigure}\centering
% \includegraphics[width=0.32\textwidth]{main_experiments_sgd/permuted_mnist_fwt.jpg}
% \includegraphics[width=0.32\linewidth]{main_experiments_sgd/split_mnist_fwt.jpg}
% \includegraphics[width=0.32\linewidth]{main_experiments_sgd/split_cifar100_fwt.jpg}
% \end{subfigure}
% \vfil
% \begin{subfigure}\centering
% \includegraphics[width=0.32\textwidth]{main_experiments_sgd/permuted_mnist_bwt.jpg}
% \includegraphics[width=0.32\linewidth]{main_experiments_sgd/split_mnist_bwt.jpg}
% \includegraphics[width=0.32\linewidth]{main_experiments_sgd/split_cifar100_bwt.jpg}
% \end{subfigure}
% % \vspace{-0.15in}
% \caption{FWI error and BWT error (FWI error on the top and BWT error on the bottom; \textbf{left:} Permuted MNIST, \textbf{middle:} Split MNIST, \textbf{right:} Split CIFAR-$100$).}
% \label{fig:wt_results}
% % \vspace{-0.2in}
% \end{figure*}

\begin{figure*}[h!]
\vspace{-0.15in}\centering
\includegraphics[width=0.32\textwidth]{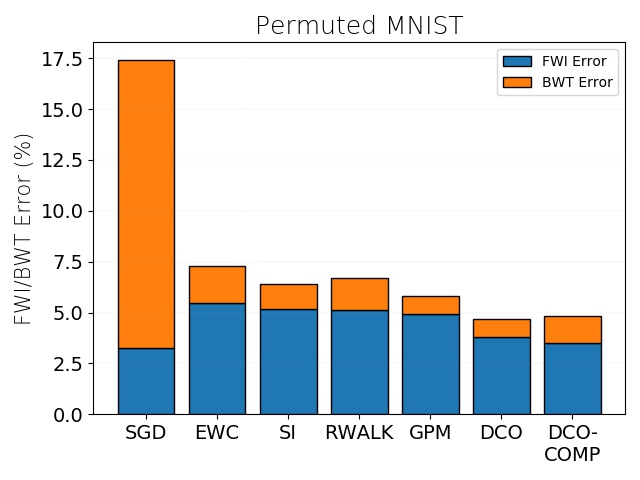}
\includegraphics[width=0.32\linewidth]{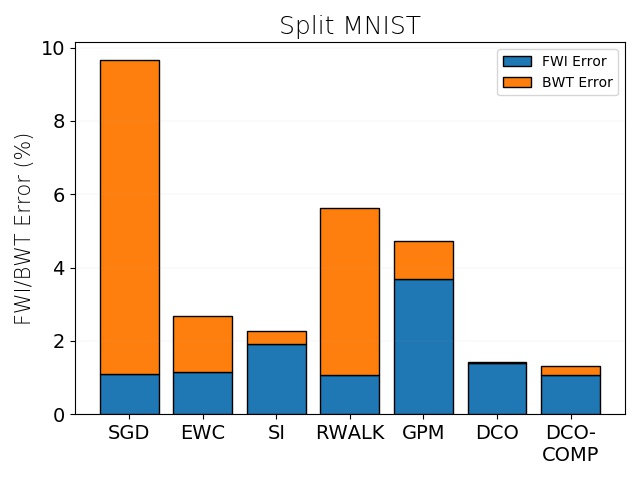}
\includegraphics[width=0.32\linewidth]{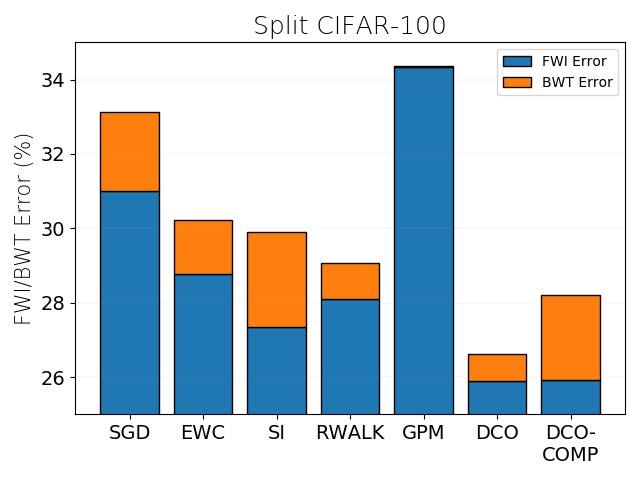}
\vspace{-0.15in}
\caption{FWI error and BWT error (\textbf{left:} Permuted MNIST, \textbf{middle:} Split MNIST, \textbf{right:} Split CIFAR-$100$).}
\label{fig:wt_results}
\vspace{-0.15in}
\end{figure*}

\begin{figure*}[h!]
\begin{minipage}{0.72\textwidth}
\includegraphics[width=0.49\textwidth]{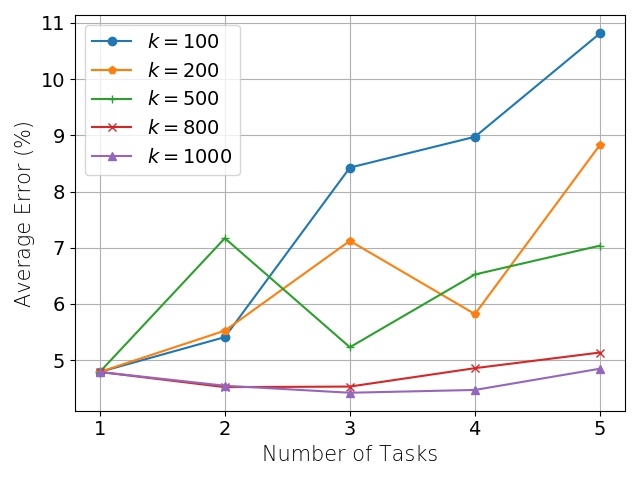}
\includegraphics[width=0.49\linewidth]{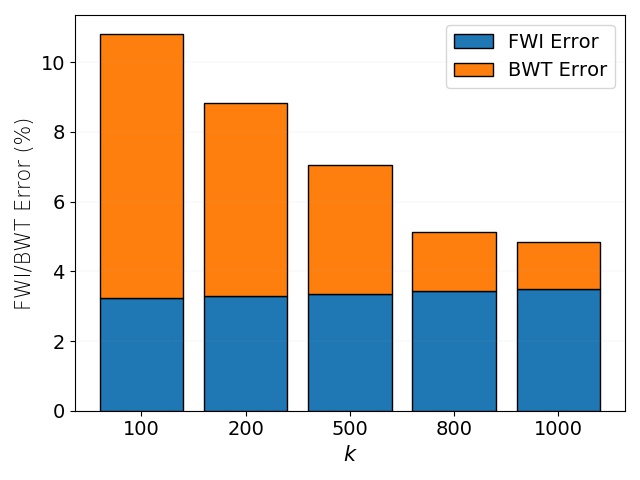}
\end{minipage}
\begin{minipage}{0.27\textwidth}
\centering
\begin{tabular}{|c||c|}
\hline
$k$ & Average Error ($\%$) \\
\hline
$100$      & $10.81$\\ \hline
$200$      & $8.83$ \\ \hline
$500$      & $7.04$ \\ \hline
$800$      & $5.14$ \\ \hline
$1000$     & $4.85$ \\\hline
\end{tabular}
\end{minipage}
\vspace{-0.16in}
\caption{DCO-COMP on permuted MNIST (\textbf{left}: Average error versus the number of tasks; \textbf{middle}: FWI and BWT errors; \textbf{right}: Final average error versus number of prohibited directions $k$).}
\label{fig:dcocomp-pmnist}
\end{figure*}

\begin{figure}[h!]
\vspace{-0.15in}
    \centering
    \includegraphics[width=0.45\linewidth]{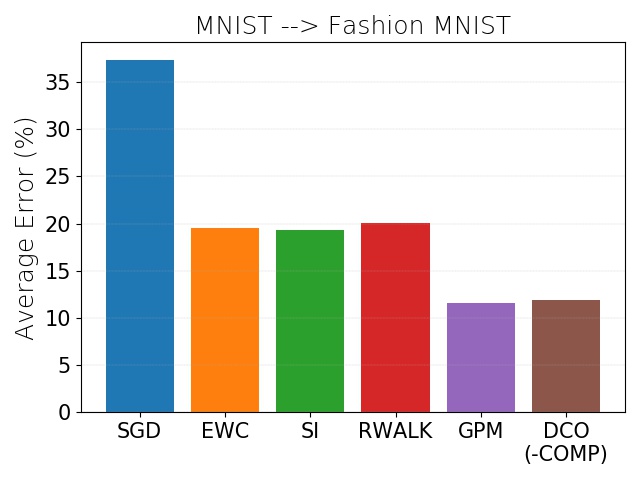}
    \vspace{-0.05in}
    \hspace{0.1in}
    \includegraphics[width=0.45\linewidth]{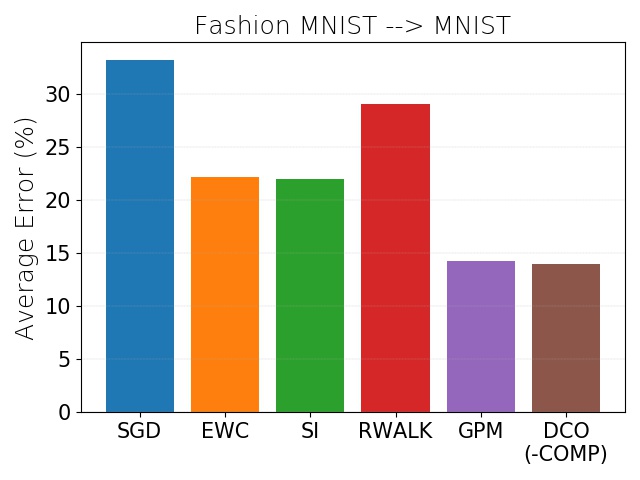}
    \vspace{-0.13in}
    \caption{Average error (\textbf{left:} MNIST $\rightarrow$ Fashion MNIST; \textbf{right:} Fashion MNIST $\rightarrow$ MNIST).}
    \label{fig:fashion_mnist_results}
    \vspace{-0.2in}
\end{figure}

% In Fig.~\ref{fig:wt_results} we report both FWI error and BWT error for each method. DCO shows strong forward-learning ability on all data sets and simultaneously suffers the least from catastrophic forgetting from among all considered techniques. Notice that although A-GEM (which is a replay strategy, not a regularization-based method) achieves the lowest FWI error on Split CIFAR-100, it has much higher BWT error than the competitor methods, including DCO.
\vspace{-0.15in}
In Fig.~\ref{fig:wt_results} we report both FWI error and BWT error for each method. In most cases DCO(-COMP) obtains the lowest FWI and BWT error among the regularization-based methods. Since the average error is the sum of FWI error and BWT error, we can conclude that DCO(-COMP) shows strong forward-learning ability while being the most efficient in alleviating the effect of catastrophic forgetting among all considered techniques.

In Fig. \ref{fig:dcocomp-pmnist} we report the memory-performance trade-off of DCO-COMP on permuted MNIST. As the number of prohibited directions $k$ increases, the FWI error slightly increases but the BWT error drops dramatically. Consequently, the DCO-COMP with largest $k$ shows the lowest final average error.

Finally, in Fig.~\ref{fig:fashion_mnist_results} we report the results of cross-domain experiment on MNIST/Fashion MNIST. DCO(-COMP) performs favorably to GPM and outperforms all other continual learning methods.

\section{Conclusion}
\label{sec:conclusion}
This paper elucidates the interplay between the local geometry of a deep learning optimization landscape and the quality of a network's performance in a continual learning setting. We derive a new continual learning algorithm counter-acting the process of catastrophic forgetting that explores the plausible manifold of parameters on which all tasks achieve good performance based on the knowledge of its geometric properties. Experiments demonstrate that this online algorithm achieves improvement in performance compared to more common approaches, which makes it a plausible method for solving a continual learning problem. Due to explicitly characterizing the manifold shared between the tasks, our work potentially provides a tool for better understanding how quickly the learning capacity of the network with a fixed architecture is consumed by adding new tasks and identifying the moment when the network lacks capacity to accommodate new coming task and thus has to be expanded. This direction will be explored in the future work.

\bibliographystyle{splncs04}
\bibliography{citation}

\begin{thebibliography}{10}
\providecommand{\url}[1]{\texttt{#1}}
\providecommand{\urlprefix}{URL }
\providecommand{\doi}[1]{https://doi.org/#1}

\bibitem{MAS}
Aljundi, R., Babiloni, F., Elhoseiny, M., Rohrbach, M., Tuytelaars, T.: Memory
  aware synapses: Learning what (not) to forget. In: ECCV (2018)

\bibitem{aljundi2017expert}
Aljundi, R., Chakravarty, P., Tuytelaars, T.: Expert gate: Lifelong learning
  with a network of experts. In: CVPR (2017)

\bibitem{aljundi2019gradient}
Aljundi, R., Lin, M., Goujaud, B., Bengio, Y.: Gradient based sample selection
  for online continual learning. In: NeurIPS (2019)

\bibitem{NEURIPS2020_4dd9cec1}
Bao, X., Lucas, J., Sachdeva, S., Grosse, R.B.: Regularized linear autoencoders
  recover the principal components, eventually. In: Larochelle, H., Ranzato,
  M., Hadsell, R., Balcan, M., Lin, H. (eds.) Advances in Neural Information
  Processing Systems. vol.~33, pp. 6971--6981 (2020)

\bibitem{bottou-98x}
Bottou, L.: Online algorithms and stochastic approximations. In: Online
  Learning and Neural Networks. Cambridge University Press (1998)

\bibitem{RWALK}
Chaudhry, A., Dokania, P.K., Ajanthan, T., Torr, P.H.S.: Riemannian walk for
  incremental learning: Understanding forgetting and intransigence. In: ECCV
  (2018)

\bibitem{chaudhry2018efficient}
Chaudhry, A., Ranzato, M.A., Rohrbach, M., Elhoseiny, M.: Efficient lifelong
  learning with a-gem. In: ICLR (2019),
  \url{https://openreview.net/forum?id=Hkf2_sC5FX}

\bibitem{chaudhry2020continual}
Chaudhry, A., Khan, N., Dokania, P.K., Torr, P.H.: Continual learning in
  low-rank orthogonal subspaces. arXiv preprint arXiv:2010.11635  (2020)

\bibitem{OGD}
Farajtabar, M., Azizan, N., Mott, A., Li, A.: Orthogonal gradient descent for
  continual learning. CoRR  \textbf{abs/1910.07104} (2019)

\bibitem{robust_evaluation_cl}
Farquhar, S., Gal, Y.: Towards robust evaluations of continual learning. CoRR
  \textbf{abs/1805.09733} (2018)

\bibitem{feng2020neural}
Feng, Y., Tu, Y.: How neural networks find generalizable solutions: Self-tuned
  annealing in deep learning. CoRR  \textbf{abs/2001.01678} (2020)

\bibitem{goodfellow2014empirical}
Goodfellow, I.J., Mirza, M., Xiao, D., Courville, A., Bengio, Y.: An empirical
  investigation of catastrophic forgetting in gradient-based neural networks.
  In: ICLR (2014)

\bibitem{He2016DeepRL}
He, K., Zhang, X., Ren, S., Sun, J.: Deep residual learning for image
  recognition. In: CVPR (2016)

\bibitem{he2019task}
He, X., Sygnowski, J., Galashov, A., Rusu, A.A., Teh, Y.W., Pascanu, R.: Task
  agnostic continual learning via meta learning. CoRR  \textbf{abs/1906.05201}
  (2019)

\bibitem{Hou2019CVPR}
Hou, S., Pan, X., Loy, C.C., Wang, Z., Lin, D.: Learning a unified classifier
  incrementally via rebalancing. In: Proceedings of the IEEE/CVF Conference on
  Computer Vision and Pattern Recognition (CVPR) (June 2019)

\bibitem{hung2019compacting}
Hung, C.Y., Tu, C.H., Wu, C.E., Chen, C.H., Chan, Y.M., Chen, C.S.: Compacting,
  picking and growing for unforgetting continual learning. In: NeurIPS (2019)

\bibitem{isele2018selective}
Isele, D., Cosgun, A.: Selective experience replay for lifelong learning. In:
  AAAI (2018)

\bibitem{EWC}
Kirkpatrick, J., Pascanu, R., Rabinowitz, N., Veness, J., Desjardins, G., Rusu,
  A.A., Milan, K., Quan, J., Ramalho, T., Grabska-Barwinska, A., et~al.:
  Overcoming catastrophic forgetting in neural networks. PNAS
  \textbf{114}(13),  3521--3526 (2017)

\bibitem{cifar}
Krizhevsky, A., Nair, V., Hinton, G.: Cifar-10 and cifar-100 datasets.
  \url{https://www. cs. toronto. edu/kriz/cifar. html} (2009)

\bibitem{NIPS2012_4824}
Krizhevsky, A., Sutskever, I., Hinton, G.E.: Imagenet classification with deep
  convolutional neural networks. In: NIPS (2012)

\bibitem{MNIST}
LeCun, Y., Boser, B., Denker, J.S., Henderson, D., Howard, R.E., Hubbard, W.,
  Jackel, L.D.: Backpropagation applied to handwritten zip code recognition.
  Neural computation  \textbf{1}(4),  541--551 (1989)

\bibitem{li2019learn}
Li, X., Zhou, Y., Wu, T., Socher, R., Xiong, C.: Learn to grow: {A} continual
  structure learning framework for overcoming catastrophic forgetting. In: ICML
  (2019), \url{http://proceedings.mlr.press/v97/li19m.html}

\bibitem{LwF}
Li, Z., Hoiem, D.: Learning without forgetting. IEEE transactions on pattern
  analysis and machine intelligence  \textbf{40}(12),  2935--2947 (2017)

\bibitem{liu2022continual}
Liu, H., Liu, H.: Continual learning with recursive gradient optimization. In:
  International Conference on Learning Representations (2022),
  \url{https://openreview.net/forum?id=7YDLgf9_zgm}

\bibitem{GEM}
Lopez-Paz, D., Ranzato, M.A.: Gradient episodic memory for continual learning.
  In: NeurIPS (2017)

\bibitem{mallya2018piggyback}
Mallya, A., Davis, D., Lazebnik, S.: Piggyback: Adapting a single network to
  multiple tasks by learning to mask weights. In: ECCV (2018)

\bibitem{mallya2018packnet}
Mallya, A., Lazebnik, S.: Packnet: Adding multiple tasks to a single network by
  iterative pruning. In: CVPR (2018)

\bibitem{mccloskey1989catastrophic}
McCloskey, M., Cohen, N.J.: Catastrophic interference in connectionist
  networks: The sequential learning problem. In: Psychology of learning and
  motivation, vol.~24, pp. 109--165. Elsevier (1989)

\bibitem{parisi2019continual}
Parisi, G., Kemker, R., Part, J., Kanan, C., Wermter, S.: Continual lifelong
  learning with neural networks: A review. Neural Networks  (2018).
  \doi{10.1016/j.neunet.2019.01.012}

\bibitem{encoder_cl}
Rannen, A., Aljundi, R., Blaschko, M.B., Tuytelaars, T.: Encoder based lifelong
  learning. In: ICCV (2017)

\bibitem{NIPS2019_8981}
Rao, D., Visin, F., Rusu, A., Pascanu, R., Teh, Y.W., Hadsell, R.: Continual
  unsupervised representation learning. In: NeurIPS (2019)

\bibitem{riemer2018learning}
Riemer, M., Cases, I., Ajemian, R., Liu, M., Rish, I., Tu, Y., Tesauro, G.:
  Learning to learn without forgetting by maximizing transfer and minimizing
  interference. In: ICLR (2019),
  \url{https://openreview.net/forum?id=B1gTShAct7}

\bibitem{ritter2018online}
Ritter, H., Botev, A., Barber, D.: Online structured laplace approximations for
  overcoming catastrophic forgetting. arXiv preprint arXiv:1805.07810  (2018)

\bibitem{rostami2019complementary}
Rostami, M., Kolouri, S., Pilly, P.K.: Complementary learning for overcoming
  catastrophic forgetting using experience replay. In: IJCAI (2019).
  \doi{10.24963/ijcai.2019/463}, \url{https://doi.org/10.24963/ijcai.2019/463}

\bibitem{rusu2016progressive}
Rusu, A.A., Rabinowitz, N.C., Desjardins, G., Soyer, H., Kirkpatrick, J.,
  Kavukcuoglu, K., Pascanu, R., Hadsell, R.: Progressive neural networks. CoRR
  \textbf{abs/1606.04671} (2016)

\bibitem{saha2021gradient}
Saha, G., Garg, I., Roy, K.: Gradient projection memory for continual learning.
  In: International Conference on Learning Representations (2021),
  \url{https://openreview.net/forum?id=3AOj0RCNC2}

\bibitem{schwarz2018progress}
Schwarz, J., Luketina, J., Czarnecki, W.M., Grabska-Barwinska, A., Teh, Y.W.,
  Pascanu, R., Hadsell, R.: Progress \& compress: A scalable framework for
  continual learning. In: ICML (2018),
  \url{http://proceedings.mlr.press/v80/schwarz18a.html}

\bibitem{shin2017continual}
Shin, H., Lee, J.K., Kim, J., Kim, J.: Continual learning with deep generative
  replay. In: NeurIPS (2017)

\bibitem{Simonyan15}
Simonyan, K., Zisserman, A.: Very deep convolutional networks for large-scale
  image recognition. In: ICLR (2015)

\bibitem{Tao2020CVPR}
Tao, X., Hong, X., Chang, X., Dong, S., Wei, X., Gong, Y.: Few-shot
  class-incremental learning. In: IEEE/CVF Conference on Computer Vision and
  Pattern Recognition (CVPR) (June 2020)

\bibitem{yoon2018lifelong}
Yoon, J., Yang, E., Lee, J., Hwang, S.J.: Lifelong learning with dynamically
  expandable networks. In: ICLR (2018),
  \url{https://openreview.net/forum?id=Sk7KsfW0-}

\bibitem{Synapses}
Zenke, F., Poole, B., Ganguli, S.: Continual learning through synaptic
  intelligence. In: ICML (2017)

\bibitem{zeno2018task}
Zeno, C., Golan, I., Hoffer, E., Soudry, D.: Task agnostic continual learning
  using online variational bayes. CoRR  \textbf{abs/1803.10123} (2018)

\end{thebibliography}

\clearpage
\newpage
%\onecolumn
\hrule height 4pt
\vskip 0.15in
\vskip -\parskip
\begin{center}
{\LARGE\bf Overcoming Catastrophic Forgetting via Direction-Constrained Optimization\\ (Supplementary Material) \par} 
\end{center}
\vskip 0.25in
\vskip -\parskip
\hrule height 1pt

% \section{Fisher information matrix}
% \label{app:fisher_information_matrix}
% We construct the following function:
% $$f(x) = f_1(x) + f_2(x) + f_3(x)$$

\section{Additional results on CIFAR-$10$ for Section~\ref{sec:observation}}
\label{app:additional_observations_cifar10}
\vspace{-0.35in}
\begin{figure}[H]
\centering
\begin{subfloat}
    \centering
    \includegraphics[width=0.7\linewidth]{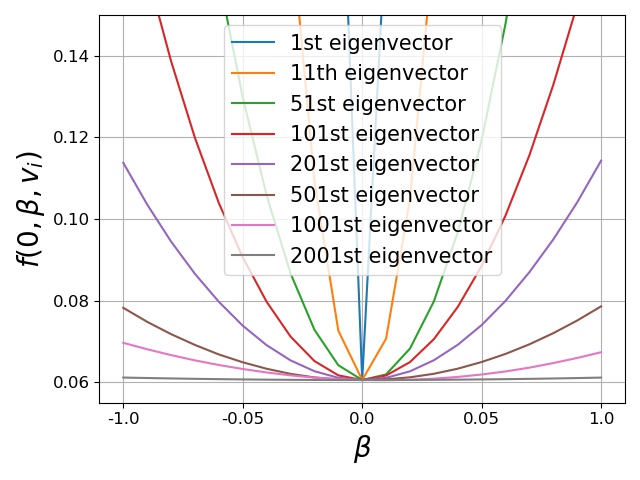}
    \vspace{-0.2in}
    \caption{The behavior of the loss function for $\alpha=0$ and varying $\beta$ when moving along different eigenvectors on CIFAR-$10$.}
    \label{fig:ob1_cifar10}
\end{subfloat}
\vspace{0.1in}
\begin{subfloat}
    \centering
    \includegraphics[width=0.7\linewidth]{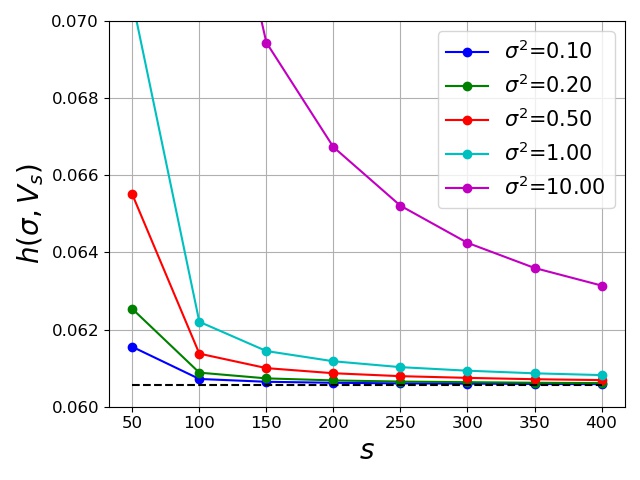}
    \vspace{-0.2in}
    \caption{The behavior of the loss function when varying $\sigma$ and $s$ on CIFAR-$10$.}
    \label{fig:ob2_cifar10}
\end{subfloat}
\end{figure}

 \begin{figure}[H]
 \vspace{-0.1in}
    \centering
    \includegraphics[width=0.45\linewidth]{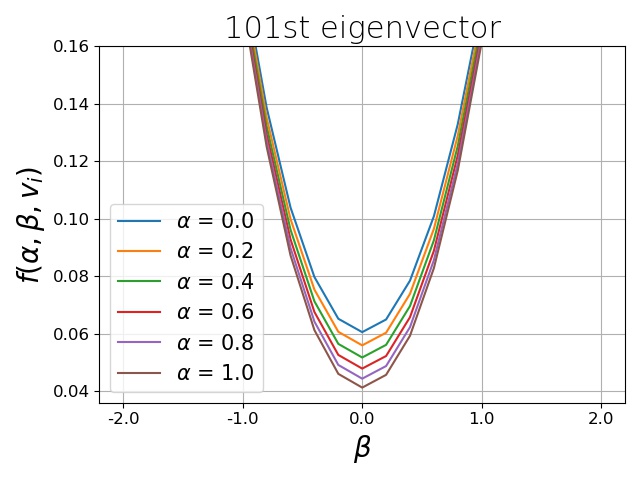}
    \includegraphics[width=0.45\linewidth]{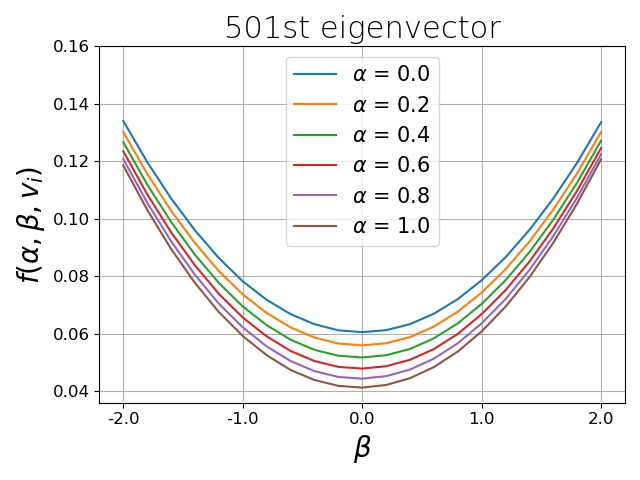}
    \includegraphics[width=0.45\linewidth]{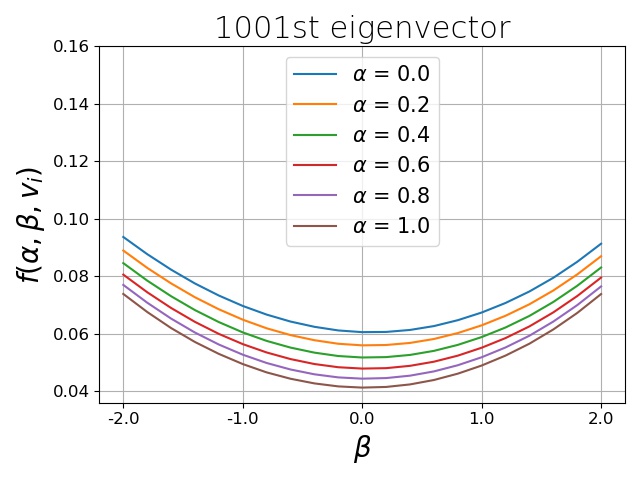}
    \includegraphics[width=0.45\linewidth]{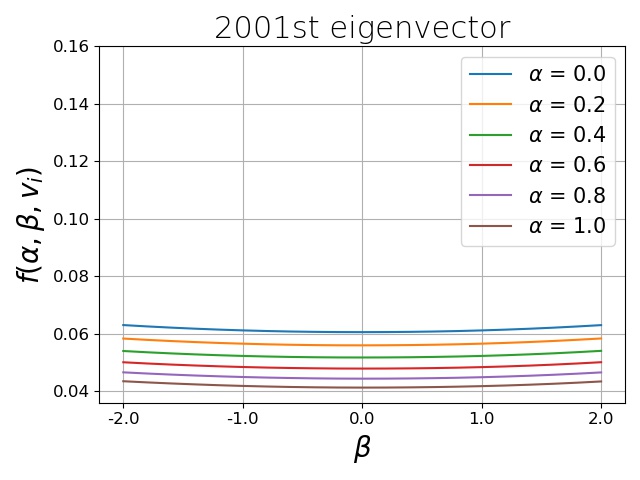}
    \vspace{-0.2in}
    \caption{The behavior of the loss function when both $\alpha$ and $\beta$ are changing for eigenvectors with different index on CIFAR-$10$.}
    \label{fig:ob3_cifar10}
    \vspace{-0.2in}
\end{figure}

\section{Difference between $GG^{\top}$ and the Fisher information matrix}
\label{app:efim}
\subsection{Theoretical difference}
DCO uses $GG^T$ and EWC uses empirical Fisher information matrix $F$ for regularization, respectively. For simplicity, we focus on a single task in isolation and explain how these two matrices are generated in different ways.

\textbf{Direction-constrained optimization (DCO)} samples the gradients of the model along the optimizer trajectory
$$GG^T = \frac{1}{N_x} \frac{1}{N_\xi} \sum_{j=1}^{N_x} \sum_{i=1}^{N_\xi} [\nabla_x L(x_j; \xi_j^i) \nabla_x L(x_j; \xi_j^i)^T] \approx \mathbb{E}_{\xi \sim p_\xi}[\nabla_x L(x; \xi) \nabla_x L(x; \xi)^T]$$

\textbf{Elastic Weight Consolidation (EWC)} samples the gradients of the model with a fixed model parameters $x^*$
$$F = \frac{1}{N_\xi}\sum_{i=1}^{N_\xi} [\nabla_x L(x^*; \xi_i) \nabla_x L(x^*; \xi_i)^T] \approx \mathbb{E}_{\xi \sim p_\xi} [\nabla_x L(x^*; \xi) \nabla_x L(x^*; \xi)^T]$$

The major difference between EWC and DCO is whether we update the parameters when we sample the gradients. EWC accumulates gradients from different mini-batches of data when model parameters are fixed to $x^*$, but DCO samples the gradients while model parameters are moving along the optimizer trajectory. More specifically, for DCO, after the model convergences we continue to train the model for another $N_x$ iterations and sample the gradients from $N_\xi$ mini-batches for each iteration.

\subsection{Illustrative example}

Consider the linear combination of three functions:
$$f(x) = 0.1* f_1(x) + 0.1* f_2(x) + 0.8* f_3(x).$$
Assume that we have no access to $f(x)$ and instead we observe $g(x)$ given as follows:
\[
  g(x) = \left.
  \begin{cases}
    % \nabla f_1(x) \quad\quad\quad \text{ with probability } 0.8 \\
    f_1(x) = 2(x-1)^2 \text{  with probability } 0.1 \\
    f_2(x) = 2(x+1)^2 \text{  with probability } 0.1 \\
    f_3(x) = 5\cdot x^2 \quad \text{ with probability } 0.8.
  \end{cases}
  \right.
\]
Note that $\mathbb{E}[g(x)] = f(x)$ and $\mathbb{E}[\nabla g(x)] = \nabla f(x)$. This setup is analogous to deep learning model training, where we are only allowed to access part of the data at each iteration and the model parameters change between iterations. 

The optimal point of function $f(x)$ is $x^*=0$ and we define $F = \frac{1}{n} \sum_{t=1}^n \nabla g(x_t)^2$, where $n$ is the number of times we sample $g(x)$. Consider two scenarios:
\begin{enumerate}
\item We fix $x_t$ and set $\forall t, x_t = x^*$: in this case $F$ becomes a Fisher information matrix
\item We start at $x_0 = x^*$ and proceed according to the update: $x_t = x_{t-1} - \eta* \nabla g(x_t)$: in this case $F$ becomes our $GG^{\top}$ matrix 
\end{enumerate}
Note that first scenario is a special case of the second one when $\eta=0$. We choose $\eta=[0.08, 0.05, 0.01, 0.0]$ and run the experiments for $10$ times. We summarize the results in the Table \ref{tab:eta}. Clearly both scenarios yield different results, i.e., with different choice of $\eta$, the values of $F$ varies.
\vspace{-0.3in}
\begin{table*}
\caption{Value of $F$ with different choice of $\eta$.}
\label{tab:eta}
\centering
\begin{tabular}{|c|c|c|c|c|}
\hline
$\eta$ &$0.08$  &$0.05$ &$0.01$ &$0.0$ \\
\hline
$F$    &$5.62$  &$4.65$ &$3.70$ &$3.53$ \\
\hline
\end{tabular}
\end{table*}
\vspace{-0.4in}

\section{Experimental details for Section~\ref{sec:observation}}
\label{app:observation_setup}
\subsection{MNIST}
We first extract images from MNIST data set \cite{MNIST} with labels of $\{0, 1\}$, then resize the original images to size $1 \times 8 \times 8$, and finally normalize each image by mean $(0.1307)$ and standard deviation $(0.3081)$.

We use a two-hidden-layer MLP to make prediction between these two classes. The numbers of neurons for each layer are ($64$-$30$-$30$-$2$) and no bias is applied. We use SGD optimizer~\cite{bottou-98x} with learning rate = $1 \times 10^{-3}$ and batch size of $128$.

We use ReLU activation function and cross-entropy loss in our experiments.
\vspace{-0.1in}

\subsection{CIFAR-$10$}
We first extract images from CIFAR-$10$ data set \cite{cifar} with labels of $\{0, 1\}$, then resize the original images to size $3 \times 8 \times 8$, and finally normalize each image by mean $(0.4914, 0.4822, 0.4465)$ and standard deviation $(0.2023, 0.1994, 0.2010)$.

We use a convolutional neural network with two convolutional layers followed by a fully-connected layer to make prediction between these two classes. Let $(ch, w, h)$ denotes the size of the input to each layer, where $ch$ is the number of input channels $w$ is the width and  $h$ is the height of the input. Let $C$ and $F$ denote convolutional layer and fully-connected layer respectively. The architecture can be described as $(3, 8, 8)C(16, 4, 4)C(32, 2, 2)F(1, 1, 2)$. There are no biases in all the layers. We use SGD optimizer~\cite{bottou-98x} with learning rate = $1 \times 10^{-2}$, momentum = $0.9$, weight decay = $10^{-4}$, and batch size of $128$.

We use ReLU activation function and cross-entropy loss in our experiments.\vspace{-0.1in}

\section{Experimental details for Section~\ref{sec:exp_res}} \vspace{-0.1in}
\subsection{Hyperparameters}
\label{app:hyperparameters}
In Table~\ref{tab:regulariaztion_hyperparameters} we summarize the setting of the regularization parameters explored for each method (except GPM which requires no regularization parameter). 
% Regarding GPM, we use its default hyperparameter setting for permuted MNIST and split CIFAR-$100$ on all MNIST experiments and CIFAR-100 experiment respectively.
When we train the linear autoencoders for DCO and DCO-COMP in step $3$ of the Algorithm~\ref{alg:continual_learning}, we always scale $L_{mse}(W;G)$ by a factor $\rho$ to avoid numerical issues.\vspace{-0.3in}

\begin{table*}[h!]
\caption{Regularizations.}
\label{tab:regulariaztion_hyperparameters}
\centering
\begin{tabular}{|c|c|c|c|c|}
 \hline
 Name & MNIST/Fashion MNIST &Permuted MNIST& Split MNIST& Split CIFAR-$100$ \\
 \hline
 EWC ($\lambda$)& $\{10^2, 10^3,  10^4,  10^5\}$ & $\{10,20 ,50, 100\}$ &  $\{10,20 ,50, 100\}$ & $\{1, 10,  10^2,  10^3\}$\\
 \hline
 SI ($c$)& $\{10^2, 10^3,  10^4,  10^5\}$ & $\{1, 10, 10^2, 10^3\}$  & $\{10^3, 10^4, 10^5, 10^6\}$ & $\{0.1, 1, 10, 10^2\}$\\
 \hline
 RWALK ($\lambda$)& $\{0.01, 0.1, 1, 10\}$ & $\{0.01, 0.1, 1, 10\}$ & $\{0.01, 0.1, 1, 10\}$ & $\{1, 2, 5, 10\}$\\
 \hline
 DCO ($\lambda$) & $\{1, 10, 100\}$ & $\{100\}$ & $\{100\}$ & $\{1000\}$\\
 \hline
 DCO-COMP ($\lambda$) & $\{1, 10, 100\}$ & $\{100\}$ & $\{100\}$ & $\{1000\}$\\
 \hline
\end{tabular}
\vspace{-0.1in}
\end{table*} \vspace{-0.6in}

% \begin{table*}[h!]
% \caption{Training settings for A-GEM.}
% \label{tab:agem_hyperparameters}
% \centering
% \begin{tabular}{|c|c|c|c|c|}
%  \hline
%  A-GEM & MNIST/Fashion MNIST & Permuted MNIST& Split MNIST& Split CIFAR-$100$ \\
%  \hline
%  Episodic Memory Size & $256$ & $256$ &  $256$ & $512$\\
%  \hline
%  Episodic Batch size & $256$ & $256$  & $256$ & $1300$\\
%  \hline
% \end{tabular}
% \vspace{-0.1in}
% \end{table*}

\begin{table*}[h!]
\caption{Training settings for DCO and DCO-COMP.}
\label{tab:dco_hyperparameters}
\centering
\begin{tabular}{|c|c|c|c|c|c|c|c|c|c|}
 \hline
 Name&$\gamma_1$ &$\gamma_2$ &$N$ &$C$ &$m$   &$\tau$ &$k$    &$\rho$ & $\theta$ \\
 \hline
 DCO&$\{0.0, 0.001\}$ &$0.1$ &$10$ &$16$ &$128$ &$2$ &$1000$ &$\{100, 1000\}$ &$\{0.0, 0.2, 0.5, 2.0\}$ \\
 \hline
 DCO-COMP&$\{0.0, 0.001\}$ &$0.1$ &$10$ &$16$ &$128$ &$2$ &$\{400, 1000, 2000\}$ &$\{100, 1000\}$ &$\{0.0, 0.2, 0.5, 2.0\}$ \\
 \hline
\end{tabular}
\vspace{-0.1in}
\end{table*}

\vspace{-0.2in}
\subsection{Autoencoder architectures}
\label{app:invauto_arch}
We use autoencoders with coupled encoder and decoder, i.e.: encoder and decoder share the same parameters. Thus we only describe encoder architectures in this section.

For each trainable layer of a classification network, we employ sets of $k$ pairs of 1-dimensional convolutional layers to process the gradients (each pair encodes separate ``prohibited'' direction). Let $L(m, n)$ denote a sequence of convolutional layers of size $1 \times m$ and size $n \times 1$ respectively. The architectures of the encoders are summarized in Table \ref{tab:autoencoder_arch}.\vspace{-0.5in}
% \textbf{MLP-100}: $L(784, 100), L(100, 100), L(100, 100), L(100, 10)$ \\
% \textbf{MLP-256}: $L(784, 256), L(256, 256), L(256, 256), L(256, 10)$ \\
% \textbf{ConvNet}: $L(27, 32), L(289, 32), L(289, 64), L(578, 64), L(1600, 100)$ \\
\begin{table}[H]
\caption{Encoder architectures. $k=1000$.}
\centering
\label{tab:autoencoder_arch}
\begin{tabular}{|c||c|} \hline
Classification & Corresponding encoder architecture \\ 
network & \\
\hline
MLP-$100$ &\makecell{\{$L(784, 100), L(100, 100), L(100, 100),$ \\ $L(100, 10)\} \times k$} \\
\hline
MLP-$256$ &\makecell{\{$L(784, 256), L(256, 256), L(256, 256),$ \\ $L(256, 10)\} \times k$} \\
\hline
ConvNet &\makecell{\{$L(27, 32), L(288, 32), L(288, 64),$ \\ $L(576, 64), L(1600, 100)\} \times k$} \\
\hline
\end{tabular}
\end{table}

% \subsection{Additional plots}
% Finally, Fig.~\ref{fig:error_zoom} provides zoomed plots for results captured in Fig.~\ref{fig:results}.
% \begin{figure*}[h!]
% \begin{subfigure}\centering 
% \includegraphics[width=0.32\textwidth]{main_experiments_sgd/permuted_mnist.jpg}\hfill
% \includegraphics[width=0.32\linewidth]{main_experiments_sgd/split_mnist.jpg}\hfill
% \includegraphics[width=0.32\linewidth]{main_experiments_sgd/split_cifar100.jpg}\hfill
% \end{subfigure}
% \begin{subfigure}\centering 
% \includegraphics[width=0.32\textwidth]{main_experiments_sgd/permuted_mnist_zoom.jpg}\hfill
% \includegraphics[width=0.32\linewidth]{main_experiments_sgd/split_mnist_zoom.jpg}\hfill
% \includegraphics[width=0.32\linewidth]{main_experiments_sgd/split_cifar100_zoom.jpg}\hfill
% \end{subfigure}
% \vspace{-0.2in}
% \caption{Average error versus the number of tasks (original plots are on the top and zoomed are on the bottom; \textbf{left:} Permuted MNIST, \textbf{middle:} Split MNIST, \textbf{right:} Split CIFAR-$100$).}
% \label{fig:error_zoom}
% \vspace{-0.15in}
% \end{figure*}

\end{document}